\documentclass{article}
\usepackage{ijcai26}

\usepackage{times}
\usepackage{soul}
\usepackage{url}
\usepackage[hidelinks]{hyperref}
\usepackage[utf8]{inputenc}
\usepackage[small]{caption}
\usepackage{graphicx}
\usepackage{amsmath}
\usepackage{amsthm}
\usepackage{booktabs}
\usepackage{algorithm}
\usepackage{algorithmic}
\usepackage[switch]{lineno}
\usepackage{booktabs}
\usepackage{multirow}
\usepackage{pifont}
\usepackage{multirow}
\usepackage{multicol}
\usepackage{arydshln}

\title{ELUTQ: Optimizing Quantization Accuracy under LUT-Based Computation for Edge LLMs}

\author{
Xin Nie$^1$
\and
Liang Dong$^1$\and
Haicheng Zhang$^{1}$\and
Jiawang Xiao$^1$ \And
Guiling Sun $^{1,*}$
\affiliations
$^1$College of Electronic Information and Optical Engineering, Nankai University\\
\emails
\{2120240458, 1120210138, zhc, 2120250407\}@mail.nankai.edu.cn, \\
sungl@nankai.edu.cn 
}

\begin{document}

\maketitle

\begin{abstract}
  Weight quantization effectively reduces memory consumption and enable the deployment of Large Language Models on edge devices, yet existing hardware-friendly methods often rely on uniform quantization, which suffers from poor weight-distribution fitting and high dequantization overhead under low-bit settings. In this paper, we propose \textbf{ELUTQ}, an efficient quantization framework featuring a novel quantization format termed \textit{Hierarchical Linear Quantization} (HLQ). HLQ is designed to better capture the statistical characteristics of weights and eliminate dequantization overhead using Bit-serial LUT-based GEMM operations. HLQ significantly improves model accuracy under low-bit settings and achieves performance comparable to QAT methods without any retraining of the weights. Moreover, an optimized quantization pipeline is integrated into ELUTQ, enabling it to complete the quantization of LLaMA 3.1–70B using only 64 GB of CPU memory and 48 GB of VRAM, reducing the hardware requirements for large-scale model quantization. To enable efficient deployment on edge devices, ELUTQ designs high-performance kernels to support end-to-end inference. Our 2-bit LLaMA3.1-8B achieves 1.5× speedup over AWQ on RTX 3090. Code is available at \url{https://github.com/Nkniexin/ELUTQ}.
\end{abstract}

\renewcommand{\thefootnote}{} 
\footnotetext{$*$ denotes the corresponding author.} 
\renewcommand{\thefootnote}{\arabic{footnote}} 

\begin{figure*}[!t]
\centering
\includegraphics[width=\linewidth]{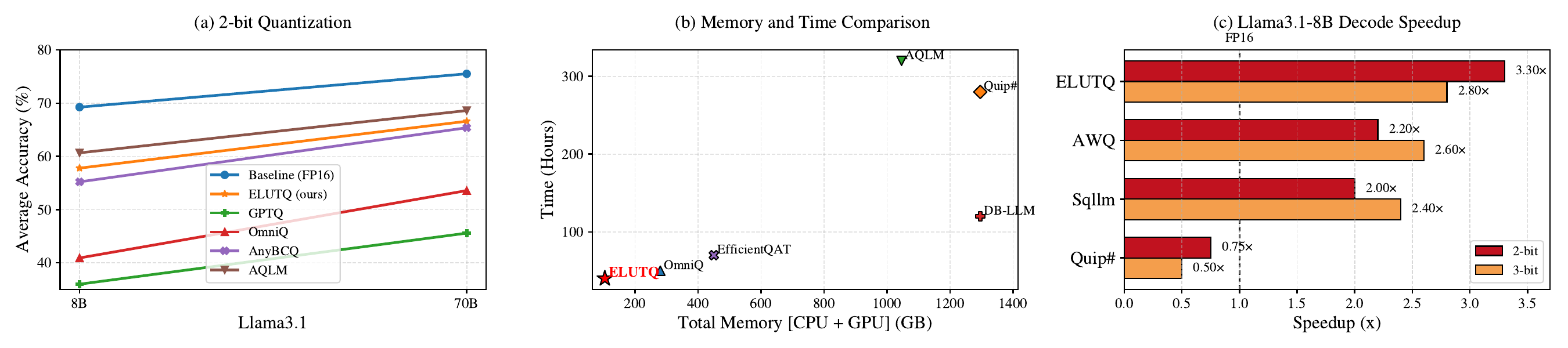} 
\caption{(a) ELUTQ outperforms uniform quantization methods but remains slightly behind codebook-based non-uniform approaches. (b) ELUTQ achieves high efficiency in both quantization time and memory usage on RTX4090s when quantizing LLaMA3.1-70B. (c) ELUTQ leverages Bit-serial LUT-based GEMM operator to significantly accelerate low-bit inference on RTX3090.}
\label{overall_results}
\end{figure*}

\section{Introduction}

Large Language Models (LLMs) have demonstrated exceptional performance across diverse tasks, including natural language understanding, image recognition, and multimodal reasoning. Traditionally deployed in cloud environments with abundant computational resources, these models are now increasingly being adapted for edge devices, such as smartphones, IoT systems, and autonomous vehicles,  to meet growing demands for low-latency inference, privacy preservation, and real-time intelligent services.

On-device hardware typically operates under stringent resource constraints, characterized by limited memory capacity and computational power. To address these challenges, model quantization~\cite{frantar2022gptq,xiao2023smoothquant,lin2024awq,kim2023squeezellm,yuan2024pbllm,DB-LLM} has emerged as a widely adopted compression technique, where high-precision weights are mapped to discrete integer values and stored using low-bit representations. This approach drastically reduces memory footprint while preserving model accuracy. Recent advances demonstrate that 8-bit weight quantization achieves near-lossless performance \cite{xiao2023smoothquant,yao2022zeroquant}. Furthermore, 4-bit quantization techniques \cite{frantar2022gptq,lin2024awq,shao2023omniquant,kim2023squeezellm,chen2024efficientqat} typically incur less than 3\% accuracy degradation, with ongoing research further improving robustness.

Quantization can be categorized as uniform \cite{frantar2022gptq,lin2024awq,shao2023omniquant,chen2024efficientqat} or non-uniform \cite{chee2023quip,10.5555/3692070.3693677,kim2023squeezellm,xu2023qalora,zhao2025ganq} depending on whether the quantization intervals are equal. Uniform quantization divides the weight space into equal intervals, making it hardware-friendly and efficient for acceleration \cite{lin2024awq,frantar2022gptq}. However, as noted in \cite{dettmers2023qlora,kim2023squeezellm}, this approach poorly matches the bell-shaped distribution of typical weights, incurring substantial approximation error. Non-uniform quantization addresses this by adaptively allocating bins, via clustering \cite{kim2023squeezellm,zhao2025ganq} or codebooks \cite{chee2023quip} to better fit the weight distribution. Although this reduces approximation error, it often sacrifices hardware compatibility due to irregular memory access patterns. To address these limitations, we propose Hierarchical Linear Quantization, a non-uniform quantization scheme that achieves a \textbf{trade-off between accuracy and hardware efficiency}.

Many large model quantization methods aim to efficiently complete model quantization, where efficiency primarily refers to time and memory consumption. Prior studies have primarily focused on reducing GPU memory consumption, the CPU memory overhead has been largely neglected. For instance, Quip\#~\cite{tseng2024quipsharp} requires at least 1,000 GB of CPU memory and over 200 hours to quantize LLaMA-3.1-70B, exceeding the capacity of most standard servers. In this paper, for HLQ, we redesign the quantization pipeline to perform quantization under strict resource constraints, maintaining low GPU memory usage, a small CPU memory footprint, and short quantization time. For example, our framework can \textbf{quantize the Llama 3.1-70B model using only 48GB VRAM and 64GB of CPU RAM in 40 hours}.

Although weight quantization reduces memory footprint, traditional methods require dequantizing low-bit weights to higher precision (e.g., INT8 or FP16) before computation. This dequantization incurs substantial unpacking overhead, which can paradoxically slow down inference at very low bit-widths. Recent advances \cite{Wei_2025} address this by replacing standard GEMM with lookup table (LUT)-based operations that implement generalized FP-INT multiplication, eliminating dequantization while achieving both linear latency reduction with bit-width and improved energy efficiency. FLGLUT \cite{park2025figlut} accelerates these operations on GPUs and T-MAC \cite{Wei_2025} optimizes for CPUs via SIMD instructions. We enhance this paradigm through a pure C++ kernel redesign that specifically supports our novel Hierarchical Linear Quantization format, maintaining cross-platform compatibility.

Our contributions are as follows.
	\begin{itemize}
  \item \textbf{Hierarchical Linear Quantization.} A novel non-uniform quantization format that provides greater flexibility in weight representation compared to uniform quantization, while preserving hardware-friendly computation.

  \item \textbf{Efficient Quantization Pipeline for HLQ.} 
  We design an efficient quantization pipeline tailored to the characteristics of the HLQ, enabling model quantization with significantly reduced memory usage and runtime.

  \item \textbf{High-Performance LUT-Based GEMM Kernels.} We design high-performance LUT–based GEMM kernels for HLQ on both CPUs and GPUs, enabling end-to-end inference across a wide range of edge devices.
\end{itemize}

\section{Related works}
\label{sec2}

\textbf{Uniform and Non-Uniform Quantization.} Uniform quantization~\cite{frantar2022gptq,lin2024awq,shao2023omniquant} is hardware-friendly, as dequantization can be implemented with simple multiplication. However, its representational capability is limited because it maps weights into a uniformly spaced range. Codebook- and clustering-based non-uniform quantization methods~\cite{tseng2024quipsharp,egiazarian2024aqlm,kim2023squeezellm,zhao2025ganq} have demonstrated strong effectiveness in modeling complex weight distributions, often yielding lower representation error than uniform quantizers. Therefore, at low bit-widths, non-uniform quantization can achieve better performance than uniform quantization. However, in modern computing systems, the smallest storage and computation unit is typically 8 bits. Therefore, for low-bit settings (e.g., 2-bit or 3-bit), the quantized integer weights must first be dequantized into 8-bit or 16-bit formats before computation, which often introduces non-negligible unpacking overhead.

\textbf{Efficient Quantization.} Efficient quantization aims to complete model quantization with minimal time and memory consumption. For example, GPTQ~\cite{frantar2022gptq} quantizes one linear layer at a time. With sufficient CPU memory, even large models such as LLaMA3.1-70B can be quantized on a single A6000 GPU. In contrast, efficient finetuning–based methods \cite{shao2023omniquant,dettmers2023spqr,dettmers2023qlora,chee2023quip,li2023loftq,xu2023qalora,DB-LLM} require partial retraining, which increases time cost but often mitigates VRAM pressure by freezing most parameters, training only a small subset, or performing quantization in batches. These strategies make quantization to be completed within limited VRAM.

\begin{figure}[t]
\centering
\includegraphics[width=\linewidth]{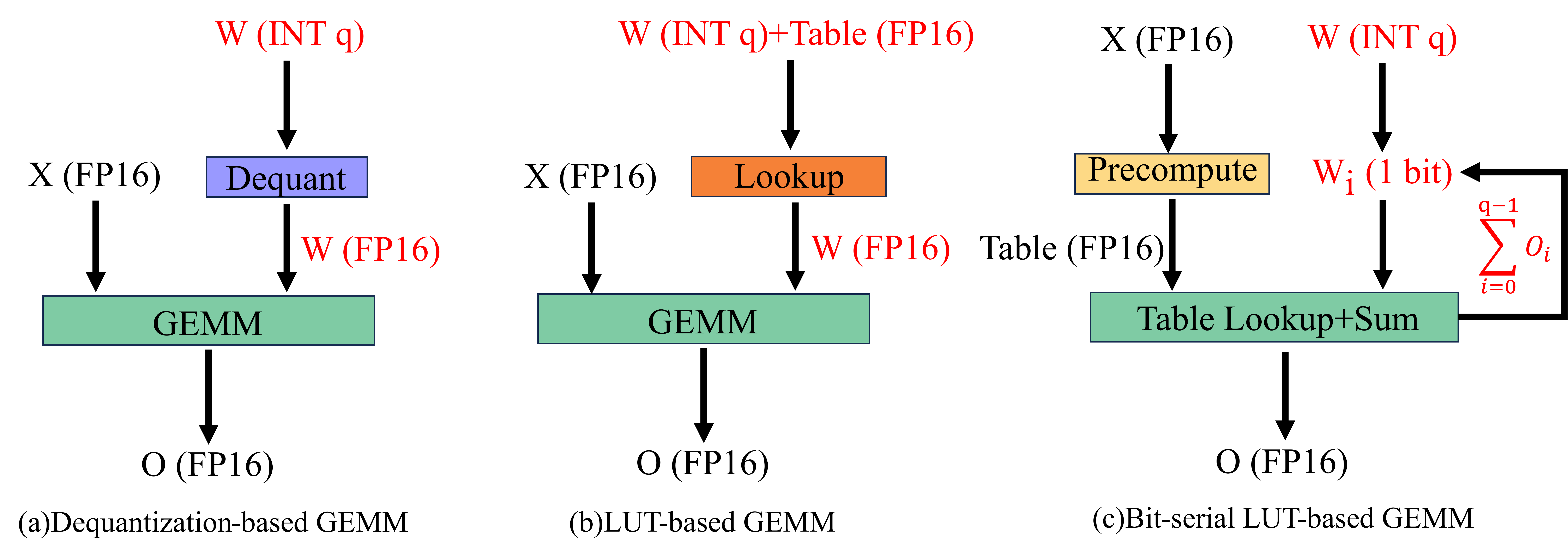} 
\caption{Illustration of two GEMM computation paradigms for weight-only quantized matrix multiplication. We provide a detailed explanations in Appendix~\ref{procedure_of_lut_gemm}.}
\label{LUT-based-GEMM}
\end{figure}

\textbf{Bit-Serial LUT-Based GEMM.} Figure~\ref{LUT-based-GEMM}(b) illustrates a GEMM paradigm termed \textbf{Bit-serial LUT-based GEMM}. Unlike the dequant-based scheme in figure~\ref{LUT-based-GEMM}(a), this method utilizes the LUT to store all possible dot products between an activation vector and single-bit weights. Then, a $q$-bit quantized weight matrix $\mathrm{W}_{int}$ is decomposed into $q$ single-bit matrices $\{\mathrm{W}_0, \mathrm{W}_1, ..., \mathrm{W}_{q-1}\}$ offline, where each element is either 0 or 1, representing the respective bit planes of the original weights. Thus, the original matrix computation requiring high-precision multiply-accumulate operations is simplified into highly efficient table lookups followed by summation (see detailed computational procedure in Appendix~\ref{procedure_of_lut_gemm}). This paradigm has been demonstrated to offer high computational efficiency and energy efficiency \cite{park2025figlut,Wei_2025}. The HLQ method is built upon the Bit-serial LUT-based GEMM computational paradigm. 


\section{Motivation}
\subsubsection{Bell-Shaped Distribution of Weights}
As noted in prior studies~\cite{dettmers2023qlora,kim2023squeezellm}, weights often exhibit a bell-shaped distribution. Uniform quantization, which maps weights to a uniformly spaced grid, struggles to accurately capture such distributions, especially under low-bit settings. Figure~\ref{figure3} shows that the weight distribution of an out\_channel in the up\_projection layer of the LLaMA3.1-8B model closely follows a Gaussian pattern, with values heavily concentrated near zero. This characteristic, commonly observed across various models, highlights the inherent limitation of uniform quantization in effectively representing natural weight distributions.

\subsubsection{Hardware Inefficiencies in Codebook and Clustering Quantization}
Methods based on codebooks (e.g., Quip\#, AQLM) or clustering (e.g., SqueezeLLM, GANQ) generally lack hardware friendliness due to irregular memory access patterns. While these approaches capture weight distributions more accurately than uniform quantization, their reliance on indirection tables or cluster lookups introduces significant computational overhead. As shown in the figure ~\ref{gpu_speed}, this overhead leads to even worse inference performance than FP16, resulting in substantial degradation in inference speed.

\subsubsection{Optimization for Bit-Serial LUT-Based GEMM}
Regarding Bit-serial LUT-Based GEMM, existing works including T-MAC \cite{Wei_2025}, LUT-GEMM \cite{park2022lut}, and FIGULT \cite{park2025figlut} have primarily focused on optimizing kernel design, yet they do not address algorithmic enhancements for improving quantization accuracy. Contemporary work like AnyBCQ~\cite{park2025anybcq} has made certain explorations, focusing on designing multi-precision models, with its 2-bit models achieving limited performance and primarily targeting model inference on GPUs. This limitations restricts the application of such kernels in broader scenarios. To bridge this gap, our work proposes HLQ to boost quantization accuracy at the algorithm level, while leveraging Bit-Serial LUT-Based GEMM to enable efficient inference on edge devices, including CPUs and GPUs.

\begin{figure}[t]
\centering
\includegraphics[width=\linewidth,]{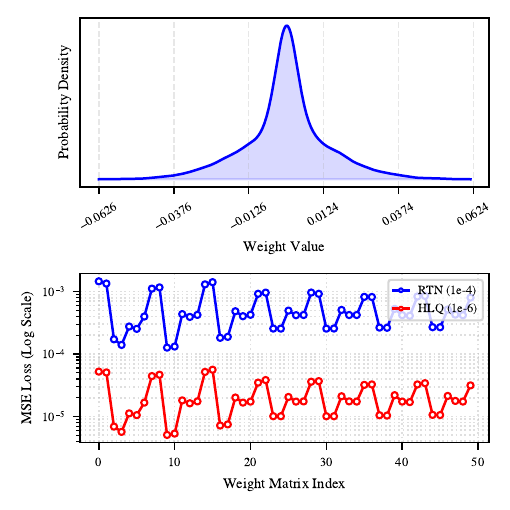} %
\caption{(Top) The weight distribution of one output channel in a up\_proj of  LLaMA3.1-8B. (Bottom) Mean squared error between quantized weights and original weights for Uniform quantization and HLQ.   }
\label{figure3}
\end{figure}
\section{Methodology}

\subsection{Hierarchical Linear Quantization}
As discussed previously, uniform quantization cannot effectively represent the distribution of weights. Inspired by Binary Coding \cite{xu2018alternating}, we propose Hierarchical Linear Quantization as an alternative approach. Specifically, for an $n$-dimensional weight vector $\mathrm{W}$, its 
$q$-bit quantized representation is denoted as $\mathrm {\hat W}$:
\begin{equation}
\label{equation3}
\mathrm{\hat W} = \sum_{j = 0}^{q-1}s_j\cdot b_j + z.
\end{equation} 
Here, $b_j$ is a binary vector $\in \{0,1\}^{n}$,$s_j\in R$ is the quantization scale, and 
$z \in R$ is the zero-point. Similar to uniform quantization, given $s$ and 
$z$, we present the quantization process of HLQ. For a $q$-bit quantization, we first generate all possible binary combinations, which form a codebook denoted by  $C \in \{0,1\}^{2^q\times q }$. Based on this codebook, we construct a candidate set:

\begin{equation}
\label{equation4}
V = s \times C^{T} + z
\end{equation} 

Then, we define \textbf{Quantization} and \textbf{Dequantization} as follows:

\textbf{Quantization:}

\begin{equation}
\label{equation5}
k^{*} = \arg\min_{k}||\mathrm{W}-V_{k}||, \ \ \ B = C_{k^{*}}
\end{equation} 

\textbf{Dequantization:}

\begin{equation}
\label{equation6}
\mathrm{\hat W} = s \times B^{T} + z
\end{equation} 

\begin{algorithm}
\caption{Alternating Optimization for Hierarchical Linear Quantization}
\label{alg:alternate_quantization}
\textbf{Input}: Weight matrix $\mathrm{W} \in R^{n \times k}$, bit width q, \\
Group size g, Max iterations $T_{max}$ \\
\textbf{Output}: scales $s \in R^{n\times \frac {k}{g} \times q}$ \\
zero points $z \in R^{n\times \frac {k}{g}}$ 
\begin{algorithmic}[1]
\STATE Reshape W into groups $\mathrm{\widetilde{W}} \in R^{n\times \frac{k}{g}\times g}  $ 
\STATE Calculate uniform quantization scale : \\
$\Delta \gets \frac {\max(\mathrm{\widetilde{W}})-\min(\mathrm{\widetilde{W}})} {2^q-1}$
\STATE Initialize  $s^{(0)} \gets [\Delta,2\Delta,...2^{q-1}\Delta]$, $z^{(0)} \gets \min(\widetilde{W})$
\STATE Generate binary combinations $C$

\FOR{$t \gets 1$ to $T_{max}$}
\STATE $\mathrm{\hat W^{(t)}} \gets \mathrm{\widetilde{W}} - z^{(t-1)}$ 

\STATE $V^{(t)} \gets s^{(t-1)} \times C^T$  \ \   \#  Matrix product




\STATE $ k^{*} \gets \arg\min_{k} \| \mathrm{\hat W^{(t)}} - V^{(t)}_k \|$\

\STATE $B^{(t)} \gets C_{k^{*}}$ \ \ \# Choose best binary combination

\STATE $s^{(t)},\ z^{(t)} \gets LSE(\mathrm{\widetilde{W}},B^{(t)})$  \# Least Squares Estimation

\ENDFOR

\end{algorithmic}
\end{algorithm}

\subsection{Weight Error Reduction via Hierarchical Linear Quantization}
\label{section_hlq}
Here, we introduce how to use Hierarchical Linear Quantization to reduce weight quantization error. Let $\mathrm{\hat W} = \mathrm{HLQ}(\mathrm{W};s,z)$,  The objective is to find the optimal set of 
$q$ scales and a zero-point $z$ to represent $\mathrm{W}$. The optimization objective can be formulated as:

\begin{equation}
\label{equation7}
\arg\min_{s,z}||\mathrm{W}-\mathrm{\hat W}||_{2}^{2}.
\end{equation}

To simplify the optimization process, we propose a heuristic alternating optimization method. At initialization, we set $z = \min(\mathrm{W})$, and the initial value of s is set to the scale factor used in uniform quantization. Specifically, let $\Delta = \frac {\max(\mathrm{W}) -\min(\mathrm{W})}{2^{q} -1} $, For a $q$-bit quantization, the initial value of s is set to $s = [\Delta,2\Delta,...2^{q-1}\Delta]$. After initialization, we solve the minimization problem in formulation~\ref{equation7} through an alternating optimization procedure between \textbf{Bit-Pattern Selection} and \textbf{Linear Reconstruction}. 

\textbf{Bit-Pattern Selection.} At $t$-th step, given the current scale parameters $s^{(t-1)}$ and zero-points $z^{(t-1)}$, we determine the optimal bit pattern $B^{(t)}$ that minimizes the reconstruction error under the fixed quantization parameters:

\begin{equation}
\label{equation_Bit_pattern}
B^{(t)}=\arg\min_B\|\mathrm{W}-\mathrm{Dequant}(B,s^{(t-1)},z^{(t-1)})\|^2,
\end{equation}
where $\mathrm{W}$ denotes original pretrained weight, $\mathrm{Dequant}$ denotes dequantization method. This process is essentially equivalent to the one described in Equation~\ref{equation5}, i.e., for each $w$, selecting the optimal bit combination from the candidate codebook.

\textbf{Linear Reconstruction.} Once the optimal bit pattern $B_t$ is obtained, we fix $B_t$ and update the continuous quantization parameters by solving a least-squares regression problem:
\begin{equation}
\label{equation_least_squares}
(s^{(t)}, z^{(t)}) = \arg\min_{s, z} \| \mathrm{W} - (B^{(t)} s + \mathbf{1}\cdot z) \|^2.
\end{equation}
Here, adding a constant column of ones allows the zero-point to be incorporated directly into the linear system. This formulation enables joint optimization of scales and zero-point given a fixed discrete structure and reduces to a standard linear least-squares problem with a closed-form solution. Therefore, the optimization objective in Equation~\ref{equation_least_squares} can be equivalently expressed as follows:
\begin{equation}
\label{LSE}
(s^{(t)}, z^{(t)}) = \mathrm{LSE}{(\mathrm{W},B^{(t)})},
\end{equation}
where $\mathrm{LSE}$ denotes the standard least squares estimation. We summarize alternating optimization in Algorithm~\ref{alg:alternate_quantization}.

Figure~\ref{figure3} shows that HLQ significantly reduces the quantization error compared to uniform quantization, with the MSE of uniform quantization at the 1e-4 level, while HLQ decreases it to the 1e-6 level. In figure~\ref{rtn_vs_hlq}, we present an ablation study showing that simply replacing RTN with HLQ, without any calibration-based fine-tuning, can substantially improve model accuracy in both the 3-bit and 2-bit settings.

It is worth noting that HLQ neither requires weight reconstruction nor introduces any additional factors, making it highly generalizable. It can be seamlessly integrated with various quantization methods.

\subsection{Efficient Finetuning for HLQ}
\begin{figure}[t]
\centering
\includegraphics[width=\linewidth]{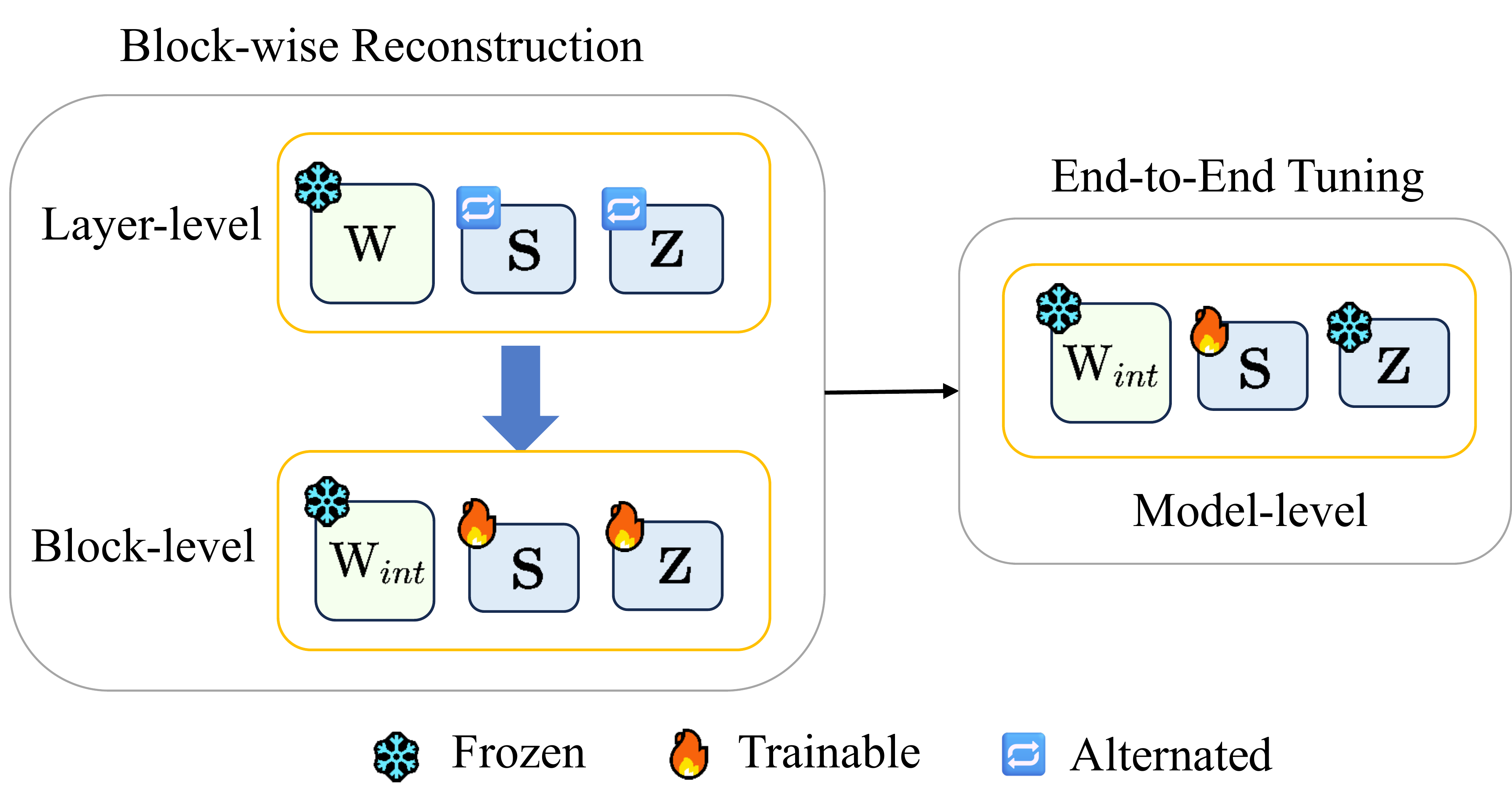} 
\caption{The pipeline of efficient finetuning for HLQ.}
\label{figure5}
\end{figure}

Previous quantization methods typically improve quantization performance by performing block-wise search for optimal quantization parameters or end-to-end finetuning. However, most of these methods are designed for uniform quantization and are not directly applicable to HLQ. As shown in figure~\ref{figure5}, We propose an efficient finetuning scheme tailored for HLQ, which consists of two stages: \textbf{block-wise reconstruction} and \textbf{end-to-end tuning}.

The block-wise reconstruction aims to minimize the block output error. We adopt the same objective but replace uniform quantization with HLQ :

\begin{equation}
\label{equation8}
\arg\min_{s,z}||\mathcal{F}(\mathbf{W},\mathbf{X})-\mathcal{F}\left(\mathrm{HLQ}(\mathbf{W};s,z),\mathbf{X}  \right) ||,
\end{equation}
where $\mathcal{F}$ denotes the mapping function of a Transformer block, 
$\mathbf{W}$ and $\mathbf{X}$ are the full-precision weights and activations, and $\mathrm{HLQ}$ represents the Hierarchical Linear Quantization function. To jointly consider local and block-level errors, we divide the block-wise reconstruction into two sub-stages. In the first sub-stage, we search for the optimal $\mathbf{W}_{int},s,z$ for each linear layer within the block using the technique described in Algorithm~\ref{alg:alternate_quantization}. In the second sub-stage, we fix $\mathbf{W}_{int}$ and only optimize the scale $s$ and zero-point $z$ by minimizing the block-level output error. For each linear layer, this strategy preserves some local information while leveraging block-level context to further refine the parameters. Additionally, this strategy is also computationally efficient. In the first sub-stage, parameter initialization for each linear layer is independent, allowing for batch-wise optimization. In the second sub-stage, with $\mathbf{W}_{int}$ fixed, there is no need to access the original full-precision weights, further reducing memory footprint.

After completing the block-wise reconstruction, we introduce end-to-end tuning, a training-based refinement approach. In this stage, we perform end-to-end optimization on the model using a calibration dataset. This technique is also adopted in EfficientQAT \cite{chen2024efficientqat}, and following their practice, we update only the quantization scale $s$ during training. 

To reduce memory consumption on both the GPU and CPU, we implemented strategies including lazy loading, CPU offloading, and disk offloading (see Appendix~\ref{system_optimization} for details). As a result, ELUTQ can quantize the LLaMA 3.1-70B model on a system configured with 64 GB of CPU RAM and 48 GB VRAM.

\subsection{Deployed on Edge Devices}
We develop high-performance CPU and GPU kernels to support GEMM, optimize the design of the lookup tables, and enable efficient end-to-end inference. Please refer to Appendix~\ref{cpu_gpu_kernels} for more details.

\section{Experiments}

We evaluate our method on several advanced pre-trained models, including LLaMA2 \cite{touvron2023llama}, LLaMA3.1 \cite{dubey2024llama3}, Qwen3 \cite{yang2025qwen3}. We use the C4 \cite{raffel2020exploring} dataset as the calibration set. Our results are compared with well-established weight-only methods, including GPTQ \cite{frantar2022gptq}, AWQ \cite{lin2024awq}, and OmniQuant \cite{shao2023omniquant}, ShiftAddLLM \cite{you2024shiftaddllm}, AnyBCQ 
\cite{park2025anybcq}, GANQ \cite{zhao2025ganq}. For 2-bit quantization, we additionally compare our method with several efficient mixed-precision quantization approaches, including PB-LLM \cite{yuan2024pbllm}, DB-LLM \cite{DB-LLM} and ApiQ \cite{liao2024apiq}. In Appendix~\ref{non_uniform_and_qat}, we also compare our method with codebook-based non-uniform quantization approaches such as Quip\#~\cite{tseng2024quipsharp} and AQLM~\cite{egiazarian2024aqlm}, as well as QAT method \cite{chen2024efficientqat}.

We evaluate all models using perplexity on the Wikitext2 \cite{wikitext2} and C4 dataset. Additionally, we assess their zero-shot capabilities using the LM Harness framework \cite{gao2021framework}. The evaluation tasks include ARC Easy, ARC Challenge \cite{clark2018think}, HellaSwag \cite{zellers2019hellaswag}, WinoGrande \cite{sakaguchi2021winogrande}, and PIQA \cite{bisk2020piqa}. For finetuning, we randomly select 4K calibration samples with sequence length of 2048 from the C4 dataset.  we set the maximum number of alternating optimization step $T_{max}$=10. In the block-wise stage, we set the learning rate to $1\times10^{-4}$ and train for 2 epochs, while in the end-to-end stage, the learning rate is set to $2\times 10^{-5}$ with 1 training epochs. All the experiments are conducted on 2 RTX4090s. For inference performance, we measure the end-to-end text generation throughput on both CPUs and GPUs using a batch size of 1.

\begin{table}[ht]
\centering
\resizebox{\linewidth}{!}{
\begin{tabular}{cccccccc}
\toprule
\textbf{Method} & \textbf{\# W}  & \textbf{\# G}&  \textbf{BPW} & \textbf{L2-7}  &  \textbf{L2-13} &  \textbf{L3.1-8} & \textbf{L3.1-70}  \\
\midrule
Baseline & 16 & - & 16   &6.97   & 6.47  & 8.89 &  6.92 \\
\midrule
GPTQ         & 2 & 128  & 2.25     & 33.70  & 20.97   & 181.82  & 22.76  \\
OmniQuant & 2   &128  & 2.25   & 15.02  &  11.05  & 31.76 & 21.79\\
ApiQ & 2 & 128 & 2.25 & \underline{12.04} & \underline{9.72} & 22.19 & - \\
PB-LLM* & - & -  & 2.2    & 20.60  & 15.32 & 57.33  & 34.52 \\
ShiftAddLLM & 2  & 128 & 2.37 &  14.60 & 9.85 &  21.97 & 18.16  \\
AnyBCQ & 2 & 128 & 2.37 & 12.17 & 9.85 & \underline{21.64} & \underline{14.15}   \\
\textbf{ELUTQ} & 2  &  128 &  2.37  & \textbf{11.21}  & \textbf{9.43} &\textbf{15.08} & \textbf{10.65}  \\ 
\midrule
DB-LLM & 2   &64 &  -  & \underline{9.62}   & \underline{8.38}  & \underline{18.74} & \underline{12.87} \\

\textbf{ELUTQ} & 2  &64 &  2.75   &  \textbf{9.12}   &  \textbf{8.02} & \textbf{14.43} & \textbf{10.21}\\

\midrule
GPTQ & 3  &128 & 3.25     &7.89  & 7.00& 11.66& 8.18  \\ 
AWQ  & 3  &128 & 3.25  & 7.84  & 6.94   & 11.58 & 8.32 \\
OmniQ& 3  &128 & 3.25  & 7.75  & 6.98 & 11.66 & 8.12  \\ 
GANQ & 3 & - & 3.25 & \underline{7.51} & \underline{6.80} & \underline{10.95} & -  \\
ShiftAddLLM & 3  & 128 & 3.50 &  7.70 & 6.92 &  11.14 & \underline{7.78}  \\
AnyBCQ & 3 & 128 & 3.50 & 7.55 & 6.84 & 12.23 & 7.62  \\
\textbf{ELUTQ} &  3& 128 & 3.50 & \textbf{7.45} & \textbf{6.77} & \textbf{10.47} & \textbf{7.42}  \\ 
\bottomrule
\end{tabular}
}
\caption{A comparison of c4 perplexity ($\downarrow$) between weight-only quantization methods, with a context length of 2048. \textbf{BPW} represents the average number of bits per weight. Scale and zero-point are assumed to be stored in fp16 format. PB-LLM* denotes the result of PB-LLM with GPTQ and 20\% salient weight. "–" indicates that the framework does not support this model. The results on Wikitext2 can be found in Appendix~\ref{full_results}.}
\label{tab:c4_ppl_results}

\end{table}

\begin{table}[ht]
\centering
\resizebox{\linewidth}{!}{
\begin{tabular}{cccccccc}
\toprule
\textbf{Method} & \textbf{\#W} & \textbf{\#G} &  L3.1-8 & L3.1-70 & Q3-8 & Q3-14 & Q3-32B \\
\midrule
FP16 & 16& - &   69.25 & 75.52 & 68.10 & 71.29 & 72.15 \\
\midrule
GPTQ & 3 & 128  & 64.21 & 71.28 & 63.04 & 67.28 & 66.44\\
AWQ & 3 & 128  & 64.98 & 72.13& 59.65 & 62.31 & 67.10\\
OmniQ& 3 & 128  & 64.46 & 71.55 & 62.76  & 64.35  & 66.89 \\
\textbf{ELUTQ} & 3 & 128  & \textbf{65.62} & \textbf{73.13} &\textbf{66.41} &\textbf{70.03} & \textbf{70.21}\\
\midrule
GPTQ & 2 & 128  & 35.98  & 45.57 & 39.45 & 48.96 & 50.34\\
OminQ & 2 & 128  & 40.91 & 53.57 & 45.17 & 53.36 & 55.28 \\
PB-LLM* & 2.2 & -  & 30.51 & 41.46 & 35.18  & 41.67 & 43.79\\
AnyBCQ & 2 & 128  & 55.21 & 65.38 & 58.32  & 61.74 & 63.81\\
\textbf{ELUTQ} & 2 & 128  & \textbf{57.79} & \textbf{66.70}&\textbf{59.44} & \textbf{64.99} & \textbf{65.57}\\

\bottomrule
\end{tabular}
}
\caption{Average accuracy (\%) on five zero-shot tasks by lm\_eval v0.4.9. PB-LLM* denotes the result of PB-LLM with GPTQ and 20\% salient weight. Full results can be found in Appendix~\ref{full_results}. Acc is reported , not acc norm. }
\label{tab:average_zeroshots_results}

\end{table}

\subsection{Main Results}

\textbf{Perplexity Results.} Table~\ref{tab:c4_ppl_results} presents the perplexity results on the C4. The results on Wikitext2 can be found in Appendix~\ref{full_results}. ELUTQ further improves model performance under low-bit quantization, particularly in the 2-bit setting. For the W2g128 configuration of Llama3.1-8B, the perplexity on C4 is further reduced to 15.08. Moreover, under the W2g64 configuration, our method surpasses DB-LLM.

\textbf{Zero-Shot Tasks.} Table~\ref{tab:average_zeroshots_results} reports the average zero-shot accuracy across five tasks. Under 3-bit quantization, ELUTQ consistently outperforms GPTQ, AWQ, and OmniQuant across all model scales, narrowing the gap to FP16. The advantage becomes more pronounced in the 2-bit setting, where ELUTQ achieves substantial gains over prior methods, including AnyBCQ. Compared with GPTQ, it achieves a roughly 90\% improvement in average accuracy on LLaMA3.1-8B.

\textbf{Compression Rates.} HLQ incurs a higher average bit-width compared to uniform quantization. As shown in table~\ref{tab::compression_rates}, for example, at 2-bit, the average bit-width is 2.25 for uniform quantization versus 2.37 for HLQ. The reason is that HLQ requires storing a separate scale for each bit plane. For LLaMA3.1-8B, HLQ at 2-bit requires an additional 0.1 GB of storage space compared to uniform quantization. However, in the context of Bit-serial LUT-based GEMM, the additional storage overhead does not translate into extra computational cost. A detailed analysis can be found in Appendix~\ref{Full_results_on_edge_devices}.

\begin{table}[ht]
\centering

\resizebox{\linewidth}{!}{
\begin{tabular}{ccccc|cc}
\toprule
\multirow{2}{*}{\textbf{Model}} & \multirow{2}{*}{\textbf{\#W}}  & \multirow{2}{*}{\textbf{\#G}} & \multicolumn{2}{c}{\textbf{Uniform}} &  \multicolumn{2}{c}{\textbf{HLQ}}  \\
 &  &  &  Rates& Mem (GB) & Rates& Mem (GB)  \\
\midrule
\multirow{2}{*}{LLaMA3.1-8B} &2&128 & 3.95&3.785 &3.84&3.887 \\
&3&128&3.25 & 4.598&3.11& 4.801 \\
\midrule
\multirow{2}{*}{LLaMA3.1-70B} &2&128 &6.00 & 21.84& 5.75&22.84 \\
&3&128&4.40 &29.81 & 4.13& 31.80  \\

\bottomrule
\end{tabular}
}
\caption{Comparison of Model Compression Rates between Uniform Quantization and Hierarchical Linear Quantization on LLaMA3.1-8B and LLaMA3.1-70B. A comprehensive presentation of the results is provided in Appendix~\ref{compression_ratios}.}
\label{tab::compression_rates}
\end{table}
\vspace{-0.3cm}
\subsection{Ablation Study}
\label{Ablation_study}
\subsubsection{HLQ vs. RTN Quantization}
\begin{figure}[t]
\centering
\includegraphics[width=\linewidth]{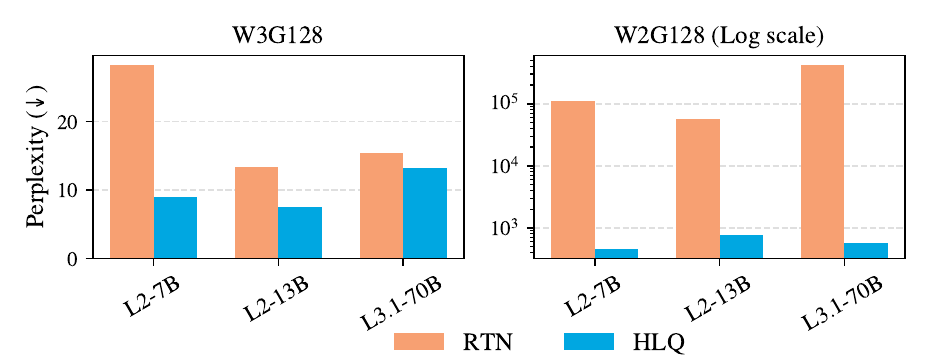} %
\caption{Comparison of HLQ and RTN quantization across multiple models under W3G128 and W2G128 settings. C4 perplexity ($\downarrow$) is reported, isolating the performance gain introduced by HLQ over standard RTN.}
\label{rtn_vs_hlq}
\end{figure}

To isolate the contribution of HLQ beyond standard RTN quantization, we do not employ any calibration-based techniques or additional optimization tricks. Instead, we simply replace RTN with HLQ. As shown in figure~\ref{rtn_vs_hlq}, this substitution consistently leads to substantial reductions in perplexity under both 3-bit and 2-bit settings, with particularly pronounced improvements in the 2-bit setting. These results indicate that HLQ can serve as a viable alternative to uniform quantization in low-bit regimes and may be integrated with other optimization techniques, such as QAT and knowledge distillation, to potentially improve performance at very low precision.

\subsubsection{Hyparameter Study for $T_{max}$}
\begin{figure}[t]
\centering
\includegraphics[width=1\linewidth]{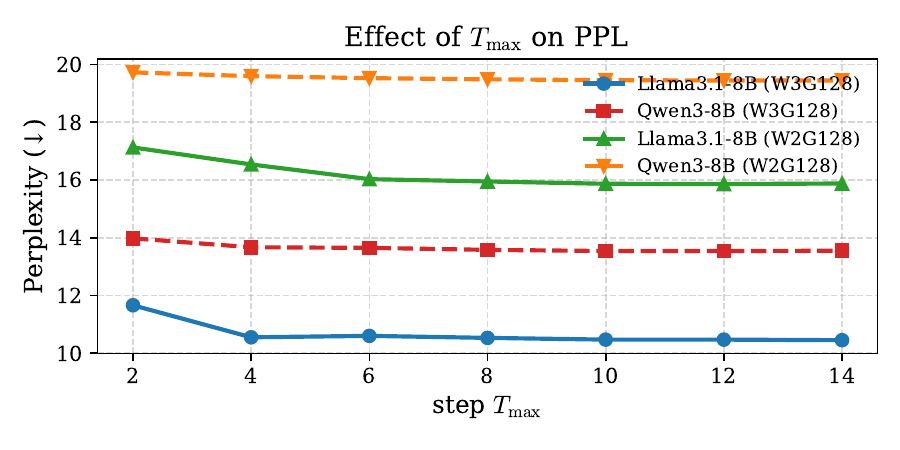} 
\caption{Impact of the alternating optimization step $T_{max}$ on quantization performance. Results are reported in terms of perplexity ($\downarrow$) on LLaMA3.1-8B and Qwen3-8B under W3G128 and W2G128 settings.}
\label{figure::Tmax_ppl_quantization}
\end{figure}

We conduct a hyparameter search experiment on the step $T_{max}$ of the alternating optimization and investigate its impact on the final model performance. The experiments are carried out on LLaMA3.1-8B and Qwen3-8B, covering both W3G128 and W2G128 settings. The results are shown in the figure~\ref{figure::Tmax_ppl_quantization}. We observe that increasing $T_{max}$ does not necessarily lead to continuous improvement. For both 3-bit and 2-bit quantization, setting $T_{max}$ = 10 yields a relatively good perplexity across both models.

\subsubsection{Impact of Each Stage in Efficient Finetuning}

Finetuning consists of two stages: block-wise quantization and end-to-end finetuning. We evaluate their individual and combined effects in Table~\ref{tab::component_influence}. Without either stage, the quantized model exhibits severe degradation. Block-wise quantization alone substantially improves both perplexity and accuracy, whereas end-to-end finetuning alone provides limited benefits. Combining both stages achieves the best performance.

\begin{table}[!ht]
\centering

\resizebox{\linewidth}{!}{
\begin{tabular}{cc|cc}
\toprule
\textbf{Block-Wise} & \textbf{E2E-Finetuning} & \textbf{C4 PPL}($\downarrow$) & \textbf{Avg.Acc($\uparrow$)}  \\
\midrule
\ding{55} & \ding{55}  & 1937.08  & 30.11 \\

\ding{51} & \ding{55}  & 17.49 &  53.78\\

\ding{55} & \ding{51} & 583.75  &  35.42 \\
\ding{51} & \ding{51} & \textbf{15.87} & \textbf{57.79} \\

\bottomrule
\end{tabular}
}
\caption{Impact of each part in finetuning on LLaMA3.1-8B W2G128 quantization.}
\label{tab::component_influence}

\end{table}

\vspace{-0.3cm}
\subsection{Efficiency Evaluation}
\label{section_efficiency_evaluation}
Table~\ref{tab::time_comparasion} summarizes the quantization time and memory consumption of several 2-bit quantization pipelines on LLaMA3.1-70B. While methods such as GPTQ and AWQ complete quantization relatively quickly, they require large amounts of CPU memory. Some finetuning approaches, including OmniQ, Quip\#, AQLM, DB-LLM, and EfficientQAT, incur substantially longer processing times ranging from tens to hundreds of hours and demand extremely high CPU memory. In comparison, ELUTQ optimizes only a small number of parameters during fine-tuning under an efficient alternating optimization algorithm and a two-stage strategy, and by incorporating lazy loading and disk offloading into the quantization pipeline, it maintains low memory requirements and completes quantization within 40 hours. We further provide a parameter efficiency evaluation in Appendix~\ref{experiment_configurations_on_memory_time}.

\begin{table}[!ht]
\centering

\resizebox{\linewidth}{!}{
\begin{tabular}{cccc}
\toprule

\textbf{Method} & \textbf{Time(h)} &\textbf{CPU Mem(GB)} &\textbf{GPU Mem(GB)} \\
\midrule
GPTQ & 1.50 & 180 & 32\\

AWQ & 2.0 & 180 & 24\\
OmniQ & 50 & 240 & 48 \\

Quip\# & 280 & 1200 & 320  \\

Aqlm & 320 & 950 &  320\\

DB-LLM & 120 & 1200 & 320 \\

EfficientQAT &  70 & 400 & 48 \\

\textbf{ELUTQ} & 40  & 64 & 48  \\

\bottomrule
\end{tabular}
}
\caption{Comparison of quantization time and peak memory usage on W2G128 Llama3.1-70B across different quantization methods. More details can be found in Appendix~\ref{experiment_configurations_on_memory_time}.}
\label{tab::time_comparasion}

\end{table}

\vspace{-0.3cm}
\subsection{Hardware Deployment on Edge Devices}
\label{hardware_deployment}
Our evaluation includes consumer-grade CPU chips such as Rk3588 and Apple M2, as well as consumer-grade GPUs including the RTX 3090 and RTX 4090. In addition, we conduct experiments on server-class accelerators, including the A100 and H100. Full results can be found in Appendix~\ref{Full_results_on_edge_devices}.

\begin{figure}[t]
\centering
\includegraphics[width=\linewidth]{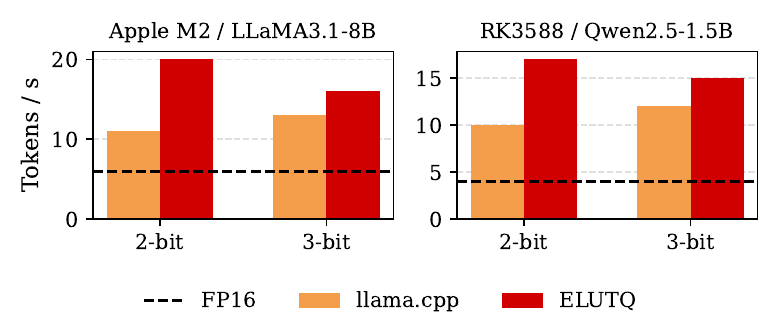} %
\caption{End to End Evaluation (tokens/s) of generating a sequence lengeth of 128 on Apple M2 and RK3588 using 4 threads. Experiments were conducted with a batch size of 1.}
\label{cpu_speed}
\end{figure}

\begin{figure}[t]
\centering
\includegraphics[width=\linewidth]{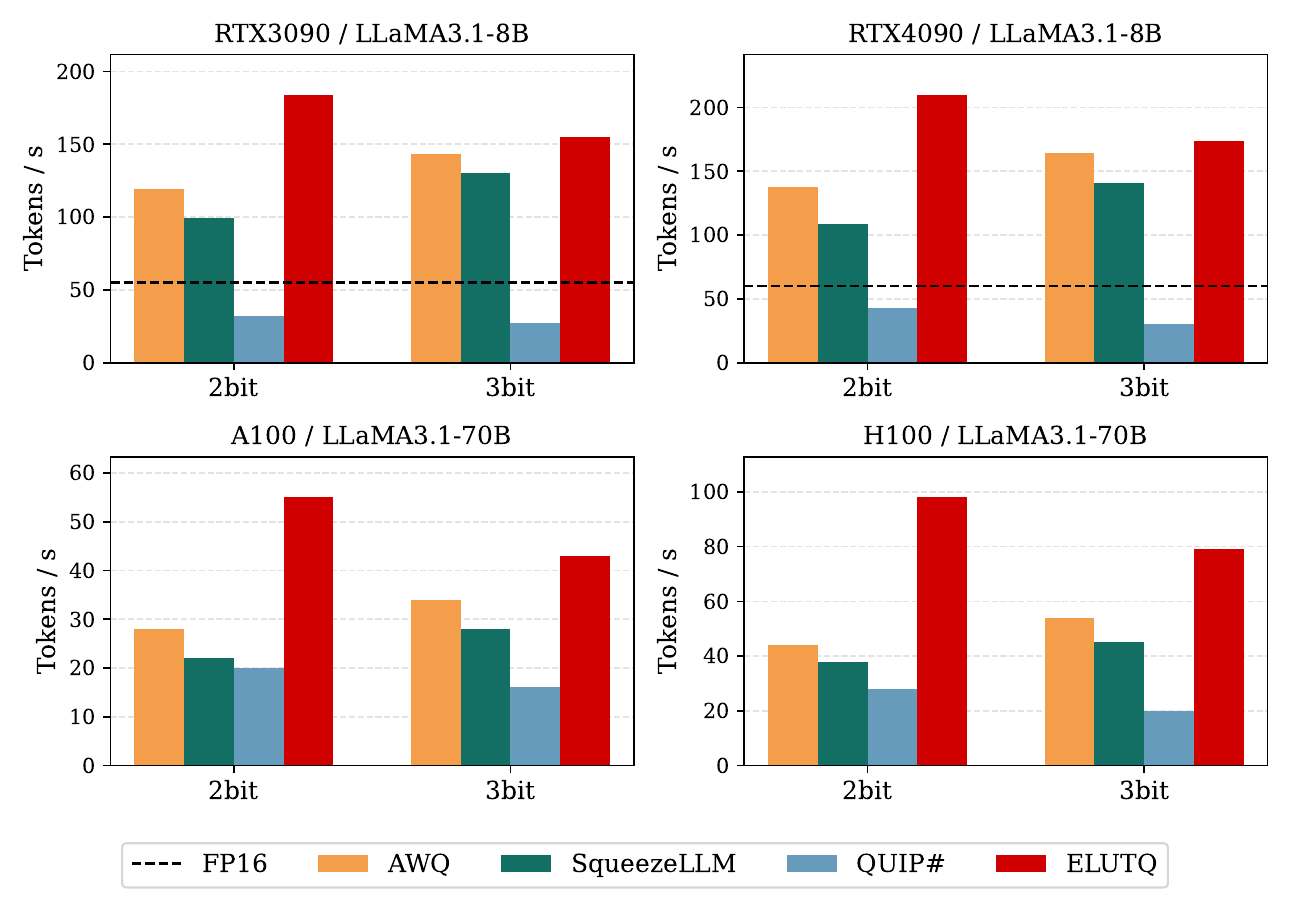} %
\caption{End to End Evaluation (tokens/s) of generating a sequence lengeth of 1024 on RTX3090, RTX4090, A100, H100. Experiments were conducted with a batch size of 1. FP16 inference of LLaMA3.1-70B is not feasible on A100 and H100 due to memory constraints.}
\label{gpu_speed}
\end{figure}

As shown in figures~\ref{cpu_speed} and~\ref{gpu_speed}, we compare our method against llama.cpp \cite{llama.cpp} for CPU inference and three representative GPU baselines, including AWQ \cite{lin2024awq} with uniform quantization, SqueezeLLM \cite{kim2023squeezellm} with LUT-based dequantization, and Quip\# \cite{tseng2024quipsharp} with codebook-based quantization, using FP16 as the baseline. On the CPUs, our 2-bit LLaMA3.1-8B model achieves an approximately 1.4× speedup over llama.cpp. On the GPUs, we observe that Quip\# performs worse than FP16 due to highly irregular memory accesses caused by codebook lookups, while SqueezeLLM, despite leveraging LUT-based dequantization, still falls behind AWQ. However, AWQ incurs increased weight unpacking overhead at lower bit-widths, resulting in inferior performance at 2-bit compared to 3-bit. In contrast, ELUTQ, built upon the bit-serial LUT-based GEMM paradigm, consistently achieves higher throughput than dequant-based methods, with inference speed further improving as the bit-width decreases. Notably, the performance gains of ELUTQ become more pronounced for larger models and lower bit-widths.

\section{Limitation and Discussion}
Here, we discuss the limitations of our work and outline several promising directions for future research.

\begin{itemize}
    \item \textbf{Weight-only quantization.} 
    The current study focuses exclusively on weight only quantization. Extending our framework to weight-activation quantization could further enhance inference efficiency and reduce memory consumption. 
    \item \textbf{Other quantization frameworks.} 
    Combining HLQ with Quantization-aware Training or LoRA-based adaptation may lead to additional gains in quantization accuracy, which we plan to explore in subsequent work.
\end{itemize}

\section{Conclusion}
In this paper, we propose \textbf{ELUTQ}, an efficient quantization framework designed for deploying large language models on edge devices. The framework is carefully aligned with the computation pipeline of Bit-serial LUT-based GEMM and introduces a novel \textit{Hierarchical Linear Quantization} method that better captures the weight distribution compared to traditional uniform quantization. Moreover, we designed an optimized quantization pipeline for HLQ to enhance model accuracy and reduce both GPU and CPU memory consumption during the quantization process. Finally, we support end-to-end inference to facilitate the practical deployment of quantized models in edge scenarios.

\newpage

\section*{Acknowledgments}
This work was supported by the National Natural Science Foundation of China (Grant No. 92473208), the Tianjin Science and Technology Planning Program (Grant No. 24ZYCGYS00680), and in part by the Tianjin Key Laboratory of Optical-electronic Sensor and Sensor Network Technology.

\bibliographystyle{named}
\bibliography{ijcai26}

@article{lin2024awq,
  title={Awq: Activation-aware weight quantization for on-device llm compression and acceleration},
  author={Lin, Ji and Tang, Jiaming and Tang, Haotian and Yang, Shang and Chen, Wei-Ming and Wang, Wei-Chen and Xiao, Guangxuan and Dang, Xingyu and Gan, Chuang and Han, Song},
  journal={Proceedings of machine learning and systems},
  volume={6},
  pages={87--100},
  year={2024}
}

@article{xu2018alternating,
  title={Alternating multi-bit quantization for recurrent neural networks},
  author={Xu, Chen and Yao, Jianqiang and Lin, Zhouchen and Ou, Wenwu and Cao, Yuanbin and Wang, Zhirong and Zha, Hongbin},
  journal={arXiv preprint arXiv:1802.00150},
  year={2018}
}

@article{chen2024efficientqat,
  title={Efficientqat: Efficient quantization-aware training for large language models},
  author={Chen, Mengzhao and Shao, Wenqi and Xu, Peng and Wang, Jiahao and Gao, Peng and Zhang, Kaipeng and Luo, Ping},
  journal={arXiv preprint arXiv:2407.11062},
  year={2024}
}

@inproceedings{park2025figlut,
  title={FIGLUT: An Energy-Efficient Accelerator Design for FP-INT GEMM Using Look-Up Tables},
  author={Park, Gunho and Kwon, Hyeokjun and Kim, Jiwoo and Bae, Jeongin and Park, Baeseong and Lee, Dongsoo and Lee, Youngjoo},
  booktitle={2025 IEEE International Symposium on High Performance Computer Architecture (HPCA)},
  pages={1098--1111},
  year={2025}
}

@article{frantar2022gptq,
  title={gptq: Accurate post-training quantization for generative pre-trained transformers},
  author={Frantar, Elias and Ashkboos, Saleh and Hoefler, Torsten and Alistarh, Dan},
  journal={arXiv preprint arXiv:2210.17323},
  year={2022}
}

@article{park2022lut,
  title={Lut-gemm: Quantized matrix multiplication based on luts for efficient inference in large-scale generative language models},
  author={Park, Gunho and Park, Baeseong and Kim, Minsub and Lee, Sungjae and Kim, Jeonghoon and Kwon, Beomseok and Kwon, Se Jung and Kim, Byeongwook and Lee, Youngjoo and Lee, Dongsoo},
  journal={arXiv preprint arXiv:2206.09557},
  year={2022}
}

@article{shao2023omniquant,
  title={Omniquant: Omnidirectionally calibrated quantization for large language models},
  author={Shao, Wenqi and Chen, Mengzhao and Zhang, Zhaoyang and Xu, Peng and Zhao, Lirui and Li, Zhiqian and Zhang, Kaipeng and Gao, Peng and Qiao, Yu and Luo, Ping},
  journal={arXiv preprint arXiv:2308.13137},
  year={2023}
}

@article{dettmers2023qlora,
  title={Qlora: Efficient finetuning of quantized llms},
  author={Dettmers, Tim and Pagnoni, Artidoro and Holtzman, Ari and Zettlemoyer, Luke},
  journal={Advances in neural information processing systems},
  volume={36},
  pages={10088--10115},
  year={2023}
}

@article{chee2023quip,
  title={Quip: 2-bit quantization of large language models with guarantees},
  author={Chee, Jerry and Cai, Yaohui and Kuleshov, Volodymyr and De Sa, Christopher M},
  journal={Advances in Neural Information Processing Systems},
  volume={36},
  pages={4396--4429},
  year={2023}
}

@article{kim2023squeezellm,
  title={Squeezellm: Dense-and-sparse quantization},
  author={Kim, Sehoon and Hooper, Coleman and Gholami, Amir and Dong, Zhen and Li, Xiuyu and Shen, Sheng and Mahoney, Michael W and Keutzer, Kurt},
  journal={arXiv preprint arXiv:2306.07629},
  year={2023}
}

@article{dettmers2023spqr,
  title={Spqr: A sparse-quantized representation for near-lossless llm weight compression},
  author={Dettmers, Tim and Svirschevski, Ruslan and Egiazarian, Vage and Kuznedelev, Denis and Frantar, Elias and Ashkboos, Saleh and Borzunov, Alexander and Hoefler, Torsten and Alistarh, Dan},
  journal={arXiv preprint arXiv:2306.03078},
  year={2023}
}

@inproceedings{Wei_2025,
   title={T-MAC: CPU Renaissance via Table Lookup for Low-Bit LLM Deployment on Edge},
   booktitle={Proceedings of the Twentieth European Conference on Computer Systems},
   author={Wei, Jianyu and Cao, Shijie and Cao, Ting and Ma, Lingxiao and Wang, Lei and Zhang, Yanyong and Yang, Mao},
   year={2025},
   pages={278–292}
}

@article{yao2022zeroquant,
  title={Zeroquant: Efficient and affordable post-training quantization for large-scale transformers},
  author={Yao, Zhewei and Yazdani Aminabadi, Reza and Zhang, Minjia and Wu, Xiaoxia and Li, Conglong and He, Yuxiong},
  journal={Advances in neural information processing systems},
  volume={35},
  pages={27168--27183},
  year={2022}
}

@inproceedings{xiao2023smoothquant,
  title={Smoothquant: Accurate and efficient post-training quantization for large language models},
  author={Xiao, Guangxuan and Lin, Ji and Seznec, Mickael and Wu, Hao and Demouth, Julien and Han, Song},
  booktitle={International conference on machine learning},
  pages={38087--38099},
  year={2023}
}

@misc{yuan2024pbllm,
      title={PB-LLM: Partially Binarized Large Language Models}, 
      author={Yuzhang Shang and Zhihang Yuan and Qiang Wu and Zhen Dong},
      year={2023},
      eprint={2310.00034},
      archivePrefix={arXiv},
      primaryClass={cs.LG},
      url={https://arxiv.org/abs/2310.00034}, 
}

@inproceedings{10.5555/3692070.3693677,
author = {Park, Yeonhong and Hyun, Jake and Cho, SangLyul and Sim, Bonggeun and Lee, Jae W.},
title = {Any-precision LLM: low-cost deployment of multiple, different-sized LLMs},
year = {2024},
booktitle = {Proceedings of the 41st International Conference on Machine Learning},
articleno = {1607},
numpages = {20},
series = {ICML'24}
}

@article{touvron2023llama,
  title={Llama 2: Open foundation and fine-tuned chat models},
  author={Touvron, Hugo and Martin, Louis and Stone, Kevin and Albert, Peter and Almahairi, Amjad and Babaei, Yasmine and Bashlykov, Nikolay and Batra, Soumya and Bhargava, Prajjwal and Bhosale, Shruti and others},
  journal={arXiv preprint arXiv:2307.09288},
  year={2023}
}

@article{raffel2020exploring,
  title={Exploring the limits of transfer learning with a unified text-to-text transformer},
  author={Raffel, Colin and Shazeer, Noam and Roberts, Adam and Lee, Katherine and Narang, Sharan and Matena, Michael and Zhou, Yanqi and Li, Wei and Liu, Peter J},
  journal={Journal of machine learning research},
  volume={21},
  number={140},
  pages={1--67},
  year={2020}
}

@inproceedings{bisk2020piqa,
  title={Piqa: Reasoning about physical commonsense in natural language},
  author={Bisk, Yonatan and Zellers, Rowan and Gao, Jianfeng and Choi, Yejin and others},
  booktitle={Proceedings of the AAAI conference on artificial intelligence},
  volume={34},
  number={05},
  pages={7432--7439},
  year={2020}
}

@article{clark2018think,
  title={Think you have solved question answering? try arc, the ai2 reasoning challenge},
  author={Clark, Peter and Cowhey, Isaac and Etzioni, Oren and Khot, Tushar and Sabharwal, Ashish and Schoenick, Carissa and Tafjord, Oyvind},
  journal={arXiv preprint arXiv:1803.05457},
  year={2018}
}

@article{sakaguchi2021winogrande,
  title={Winogrande: An adversarial winograd schema challenge at scale},
  author={Sakaguchi, Keisuke and Bras, Ronan Le and Bhagavatula, Chandra and Choi, Yejin},
  journal={Communications of the ACM},
  volume={64},
  number={9},
  pages={99--106},
  year={2021}
}

@article{zellers2019hellaswag,
  title={Hellaswag: Can a machine really finish your sentence?},
  author={Zellers, Rowan and Holtzman, Ari and Bisk, Yonatan and Farhadi, Ali and Choi, Yejin},
  journal={arXiv preprint arXiv:1905.07830},
  year={2019}
}

@article{DB-LLM,
  title={Db-llm: Accurate dual-binarization for efficient llms},
  author={Chen, Hong and Lv, Chengtao and Ding, Liang and Qin, Haotong and Zhou, Xiabin and Ding, Yifu and Liu, Xuebo and Zhang, Min and Guo, Jinyang and Liu, Xianglong and others},
  journal={arXiv preprint arXiv:2402.11960},
  year={2024}
}

@article{dubey2024llama3,
  title={The llama 3 herd of models},
  author={Dubey, Abhimanyu and Jauhri, Abhinav and Pandey, Abhinav and Kadian, Abhishek and Al-Dahle, Ahmad and Letman, Aiesha and Mathur, Akhil and Schelten, Alan and Yang, Amy and Fan, Angela and others},
  journal={arXiv e-prints},
  pages={arXiv--2407},
  year={2024}
}

@article{wikitext2,
  title={Pointer sentinel mixture models},
  author={Merity, Stephen and Xiong, Caiming and Bradbury, James and Socher, Richard},
  journal={arXiv preprint arXiv:1609.07843},
  year={2016}
}

@article{zhao2025ganq,
  title={GANQ: GPU-Adaptive Non-Uniform Quantization for Large Language Models},
  author={Zhao, Pengxiang and Yuan, Xiaoming},
  journal={arXiv preprint arXiv:2501.12956},
  year={2025}
}

@article{gao2021framework,
  title={A framework for few-shot language model evaluation},
  author={Gao, Leo and Tow, Jonathan and Biderman, Stella and Black, Shawn and DiPofi, Anthony and Foster, Charles and Golding, Laurence and Hsu, Jasmine and McDonell, Kyle and Muennighoff, Niklas and others},
  journal={Version v0. 0.1. Sept},
  volume={10},
  pages={8--9},
  year={2021}
}

@misc{llama.cpp,
  author       = {Gerganov, Georgi},
  title        = {llama.cpp},
  year         = {2023},
  howpublished = {\url{https://github.com/ggml-org/llama.cpp}},
  note         = {Accessed: July 31, 2025}
}

@article{yang2025qwen3,
  title={Qwen3 technical report},
  author={Yang, An and Li, Anfeng and Yang, Baosong and Zhang, Beichen and Hui, Binyuan and Zheng, Bo and Yu, Bowen and Gao, Chang and Huang, Chengen and Lv, Chenxu and others},
  journal={arXiv preprint arXiv:2505.09388},
  year={2025}
}

@article{li2023loftq,
  title={Loftq: Lora-fine-tuning-aware quantization for large language models},
  author={Li, Yixiao and Yu, Yifan and Liang, Chen and He, Pengcheng and Karampatziakis, Nikos and Chen, Weizhu and Zhao, Tuo},
  journal={arXiv preprint arXiv:2310.08659},
  year={2023}
}

@article{xu2023qalora,
  title={Qa-lora: Quantization-aware low-rank adaptation of large language models},
  author={Xu, Yuhui and Xie, Lingxi and Gu, Xiaotao and Chen, Xin and Chang, Heng and Zhang, Hengheng and Chen, Zhengsu and Zhang, Xiaopeng and Tian, Qi},
  journal={arXiv preprint arXiv:2309.14717},
  year={2023}
}

@inproceedings{liu2022ste,
  title={Nonuniform-to-uniform quantization: Towards accurate quantization via generalized straight-through estimation},
  author={Liu, Zechun and Cheng, Kwang-Ting and Huang, Dong and Xing, Eric P and Shen, Zhiqiang},
  booktitle={Proceedings of the IEEE/CVF conference on computer vision and pattern recognition},
  pages={4942--4952},
  year={2022}
}

@inproceedings{wolf-etal-2020-transformers,
    title = "Transformers: State-of-the-Art Natural Language Processing",
    author = "Thomas Wolf and Lysandre Debut and Victor Sanh and Julien Chaumond and Clement Delangue and Anthony Moi and Pierric Cistac and Tim Rault and Rémi Louf and Morgan Funtowicz and Joe Davison and Sam Shleifer and Patrick von Platen and Clara Ma and Yacine Jernite and Julien Plu and Canwen Xu and Teven Le Scao and Sylvain Gugger and Mariama Drame and Quentin Lhoest and Alexander M. Rush",
    booktitle = "Proceedings of the 2020 Conference on Empirical Methods in Natural Language Processing: System Demonstrations",
    month = oct,
    year = "2020",
    address = "Online",
    publisher = "Association for Computational Linguistics",
    url = "https://www.aclweb.org/anthology/2020.emnlp-demos.6",
    pages = "38--45"
}

@article{park2025anybcq,
  title={AnyBCQ: Hardware Efficient Flexible Binary-Coded Quantization for Multi-Precision LLMs},
  author={Park, Gunho and Bae, Jeongin and Kwon, Beomseok and Kim, Byeongwook and Kwon, Se Jung and Lee, Dongsoo},
  journal={arXiv preprint arXiv:2510.10467},
  year={2025}
}

@article{you2024shiftaddllm,
  title={Shiftaddllm: Accelerating pretrained llms via post-training multiplication-less reparameterization},
  author={You, Haoran and Guo, Yipin and Fu, Yichao and Zhou, Wei and Shi, Huihong and Zhang, Xiaofan and Kundu, Souvik and Yazdanbakhsh, Amir and Lin, Yingyan Celine},
  journal={Advances in Neural Information Processing Systems},
  volume={37},
  pages={24822--24848},
  year={2024}
}

@article{liao2024apiq,
  title={Apiq: Finetuning of 2-bit quantized large language model},
  author={Liao, Baohao and Herold, Christian and Khadivi, Shahram and Monz, Christof},
  journal={arXiv preprint arXiv:2402.05147},
  year={2024}
}

@article{tseng2024quipsharp,
  title={Quip\#: Even better llm quantization with hadamard incoherence and lattice codebooks},
  author={Tseng, Albert and Chee, Jerry and Sun, Qingyao and Kuleshov, Volodymyr and De Sa, Christopher},
  journal={arXiv preprint arXiv:2402.04396},
  year={2024}
}

@article{egiazarian2024aqlm,
  title={Extreme compression of large language models via additive quantization},
  author={Egiazarian, Vage and Panferov, Andrei and Kuznedelev, Denis and Frantar, Elias and Babenko, Artem and Alistarh, Dan},
  journal={arXiv preprint arXiv:2401.06118},
  year={2024}
}

\appendix
\onecolumn

\section*{Appendix}

\addcontentsline{toc}{section}{Appendix}

\begin{center}
\textbf{Appendix Overview}
\end{center}

\noindent
\textbf{Appendix~\ref{gradient_based_methods}:Gradient-Based Search for Hierarchical Linear Quantization} \\
We introduce an alternative gradient-based approach for searching the scale and zero-point of HLQ.

\vspace{0.5em}
\noindent
\textbf{Appendix~\ref{procedure_of_lut_gemm}: Detailed Computational procedure for Bit-Serial Lut-Based GEMM} \\
We provide a detailed description of the computation procedure for Bit-Serial LUT-based GEMM.

\vspace{0.5em}
\noindent
\textbf{Appendix~\ref{system_optimization}: System-level Optimizations for the Quantization Pipeline} \\
We present key technical components, including lazy loading, chunked computation, and disk offloading, to enable model quantization for large-scale models under limited CPU and GPU memory budgets.

\vspace{0.5em}
\noindent
\textbf{Appendix~\ref{cpu_gpu_kernels}: Design of High Performance CPU and GPU kernels} \\
Core design details of the GPU and CPU kernels.

\vspace{0.5em}
\noindent
\textbf{Appendix~\ref{compression_ratios}: Compression Ratios and Actual Memory Footprint of Quantized Models} \\
Details of Compression Ratios and Actual Memory Footprint.

\vspace{0.5em}
\noindent
\textbf{Appendix~\ref{non_uniform_and_qat}: Compared with Non-Uniform and Quantization-Aware Training Methods} \\
Further comparison of ELUTQ with codebook-based non-uniform quantization methods and QAT approaches.

\vspace{0.5em}
\noindent
\textbf{Appendix~\ref{experiment_configurations_on_memory_time}: More details of Efficiency Evaluation} \\
Dtailed experimental configurations used to measure the computational memory consumption.

\vspace{0.5em}
\noindent
\textbf{Appendix~\ref{Full_results_on_edge_devices}: Full Results of Inference Performance on Edge Devices} \\
We comprehensively present the inference performance of ELUTQ on both CPU and GPU, including kernel-level evaluation and end-to-end evaluation.

\vspace{0.5em}
\noindent
\textbf{Appendix~\ref{full_results}: Full Results on Model Accuracy} \\
We present full results on model accuracy.

\section{Gradient-Based Search for Hierarchical Linear Quantization}
\label{gradient_based_methods}
The gradient-based search updates the quantization parameters by directly computing the MSE and propagating the gradients. At initialization, we set $z = \min(\mathrm{W})$, and the initial value of s is set to the scale factor used in uniform quantization. Specifically, let $\Delta = \frac {\max(\mathrm{W}) -\min(\mathrm{W})}{2^{q} -1} $, For a $q$-bit quantization, the initial value of s is set to $s = [\Delta,2\Delta,...2^{q-1}\Delta]$. The workflow of the search process is illustrated in Algorithm~\ref{alg:gradient_based_quantization}. Here, we adopt an approximately differentiable optimization strategy to jointly update the scales and zero-points. The core idea is to decouple the quantization process into two stages: discrete selection and continuous reconstruction, thereby circumventing the gradient issues inherent in discrete operations.
\begin{itemize}
\item \textbf{Forward Pass}: the optimal combination of discrete codewords is determined via nearest-neighbor search. Subsequently, this combination is used together with learnable scaling factors and zero-points to reconstruct the quantized weights through a continuous weighted summation.
\item \textbf{Backward Pass}: During backpropagation, gradients cannot flow through the argmin-based discrete selection step. However, they can naturally propagate back through the continuous reconstruction step to update the scaling factors and zero-points. This enables the continuous parameters to be effectively optimized via standard gradient descent.
\end{itemize}
To ensure training stability, we enforce a non-negativity constraint on the scaling factors and apply dynamic range clipping to the zero-point values, which effectively prevents training divergence.
From an optimization perspective, this approach is conceptually aligned with the straight-through estimator (STE)~\cite{liu2022ste} philosophy. However, a key distinction lies in the fact that instead of relying on handcrafted gradient approximation rules, we carefully design the quantization expression itself to allow gradients to flow naturally to the continuous parameters, resulting in a more elegant approximation of gradient flow.

\begin{algorithm}
\caption{Gradient-Based Optimization for Hierarchical Linear Quantization}
\label{alg:gradient_based_quantization}
\begin{algorithmic}[1]
\REQUIRE Weight matrix $\mathrm{W} \in R^{n \times k}$, bit width q, \\
Group size g, Max iterations $T_{max}$, Tolerance $\varepsilon$
\ENSURE  scales $s \in R^{n\times \frac {k}{g} \times q}$, zero points $z \in R^{n\times \frac {k}{g}}$

\STATE Reshape W into groups $\mathrm{\widetilde{W}} \in R^{n\times \frac{k}{g}\times g}  $ 
\STATE Calculate uniform quantization scale : \\
$\Delta \gets \frac {\max(\mathrm{\widetilde{W}})-\min(\mathrm{\widetilde{W}})} {2^q-1}$
\STATE Initialize  $s^{(0)} \gets [\Delta,2\Delta,...2^{q-1}\Delta]$, $z^{(0)} \gets \min(\widetilde{W})$, $preL \gets INF$
\STATE Generate binary combinations $C$

\FOR{$t \gets 1$ to $T_{max}$}
\STATE $\mathrm{\hat W^{(t)}} \gets \mathrm{\widetilde{W}} - z^{(t-1)}$ 

\STATE $V^{(t)} \gets s^{(t-1)} \times C^T$  \ \   \#  Matrix product

\STATE $ k^{*} \gets \arg\min_{k} \| \mathrm{\hat W^{(t)}} - V^{(t)}_k \|$\

\STATE $B^{(t)} \gets C_{k^{*}}$ \ \ \# Choose best binary combination

\STATE $\mathrm{\hat W^{(t)}} \gets s^{(t-1)} \times {B^{(t)}}^T$

\STATE $\mathrm{\hat W^{(t)}} \gets \mathrm{\hat W^{(t)}} + z^{(t-1)}$

\STATE $L \gets mean((\mathrm{\widetilde{W}} -\mathrm{\hat W^{(t)}} )^2)$

\STATE $s^{(t)},\ z^{(t)} \gets Backpropagate(L)$ 

\IF{$L-preL < \varepsilon $} 
\STATE break
\ENDIF
\STATE $perL \gets L$
\ENDFOR

\end{algorithmic}
\end{algorithm}

\newpage
\section{Detailed Computational procedure for Bit-Serial Lut-Based GEMM }
\label{procedure_of_lut_gemm}
\begin{figure*}[t]\rmfamily
\centering
\includegraphics[width=\linewidth]{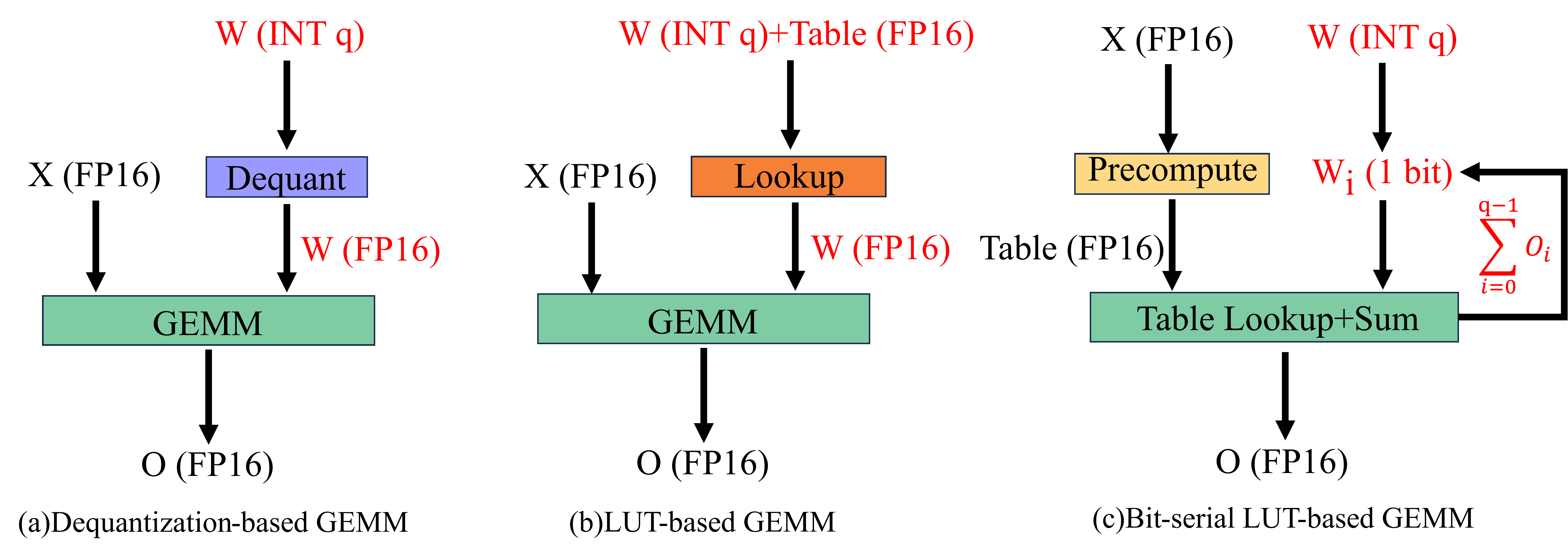} 
\caption{Illustration of three computation paradigms for weight-only quantized matrix multiplication.}
\label{three_gemms}
\end{figure*}

Here, we provide a detail computational procedure for Bit-serial Lut-based GEMM. There are two primary computational paradigms for LUT-based GEMM, as illustrated in the Figure~\ref{three_gemms}.

\begin{figure*}[t]\rmfamily
\centering
\includegraphics[width=\linewidth]{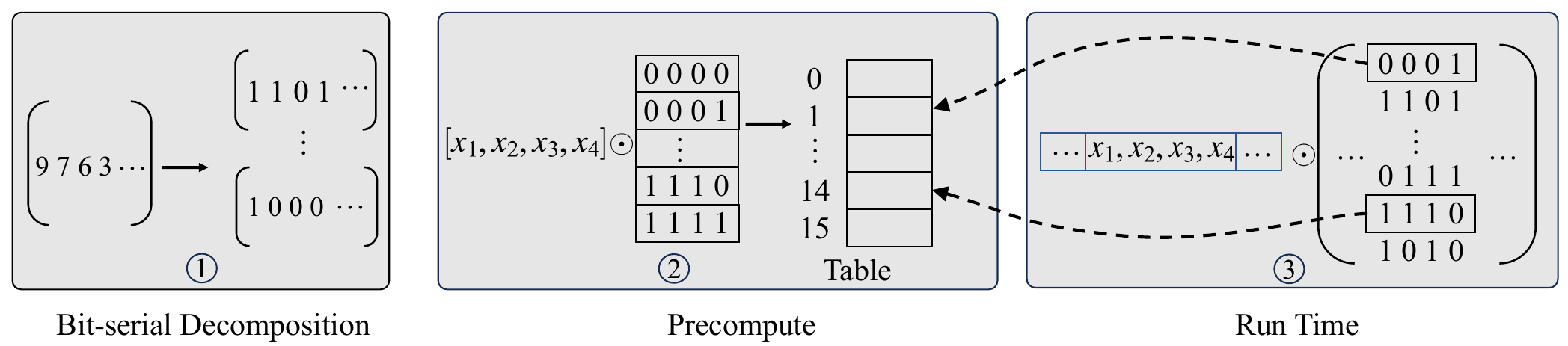} 
\caption{Detail pipeline of Bit-serial Lut-based GEMM.}
\label{bit_serial}
\end{figure*}

Figure~\ref{three_gemms}(b) depicts the first implementation scheme of LUT-based GEMM. Its core idea is to directly store the high-precision weight values corresponding to low-bit integer weights in the LUT. Consequently, during dequantization, high-precision weights can be reconstructed via direct LUT accesses instead of computationally expensive arithmetic operations, significantly reducing the overhead of dequantization. It should be noted that, in this paradigm, although weights are reconstructed into a high-precision format via the LUT, they maintain the same numerical precision (e.g., both FP16) as the activations during the matrix multiplication.

Figure~\ref{three_gemms}(c) illustrates another GEMM paradigm termed \textbf{Bit-serial LUT-based GEMM}. Unlike the scheme in Figure 2(b), this method utilizes the LUT to store all possible dot products between an activation vector and single-bit weights. The detail computational procedure is shown in Figure~\ref{bit_serial}. First, a $q$-bit quantized weight matrix $\mathrm{W}_{int}$ is decomposed into $q$ single-bit matrices $\{W_0, W_1, ..., W_{q-1}\}$ offline, where each element is either 0 or 1, representing the respective bit planes of the original weights. For example, for integer values (9, 7, 6, 3) with binary representations (1001, 0111, 0110, 0011), the matrix for the lowest bit is (1, 1, 0, 1), and the matrix for the highest bit is (1, 0, 0, 0). This decomposition is performed offline, incurring no runtime overhead. During inference, for an activation vector of the group size $g$, the system precomputes the dot products between this activation vector and all $2^g$ possible combinations of single-bit weights, storing the results in the LUT. Thus, the original matrix computation requiring high-precision multiply-accumulate operations is simplified into highly efficient table lookups followed by summation. This paradigm has been demonstrated to offer high computational efficiency and energy efficiency \cite{park2025figlut,Wei_2025}. For example, FIGLUT \cite{park2025figlut} optimized the table structure for GPU architectures to avoid bank conflicts, while T-MAC leveraged CPU vectorized lookup instructions (AVX2/NEON) to enable efficient LUT operations on CPUs.
\newpage

\section{System-level Optimizations for the Quantization Pipeline}
\label{system_optimization}
\begin{figure*}[t]
\centering
\includegraphics[width=\linewidth]{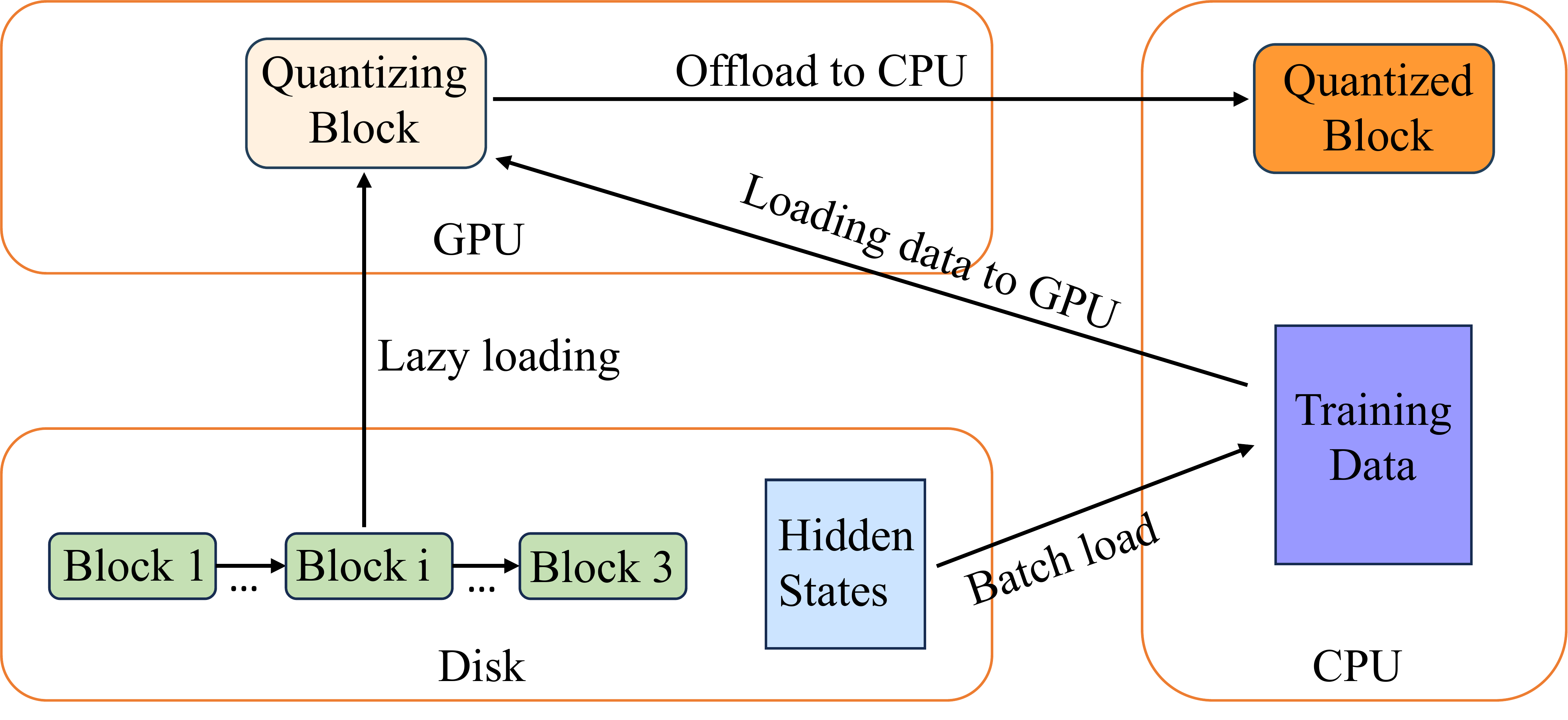} 
\caption{overview of system-level optimizations for the quantization pipeline.}
\label{overview_of_system_optimization}
\end{figure*}

As illustrated in the figure~\ref{overview_of_system_optimization} , we present how to perform model quantization for large-scale models with limited CPU and GPU memory.
\subsection{Lazy Loading}
In both GPTQ’s Hessian-based quantization and Efficient Finetuning’s block reconstruction, the quantization unit is a Transformer block. Quantizing a single block does not depend on other blocks, neither the already quantized nor the unquantized ones. Therefore, instead of loading the entire model into memory, we can load only one transformer block at a time from disk. Thanks to the sharded weight storage mechanism in the Transformers \cite{wolf-etal-2020-transformers} framework, we can leverage the weight mapping file (e.g., \texttt{model.safetensors.index.json}) to selectively load a specific block from disk for quantization. Once the quantization of a block on the GPU is completed, it is immediately offloaded to the CPU to reduce GPU memory usage.

\begin{figure}[t]
\centering
\includegraphics[width=0.5\linewidth]{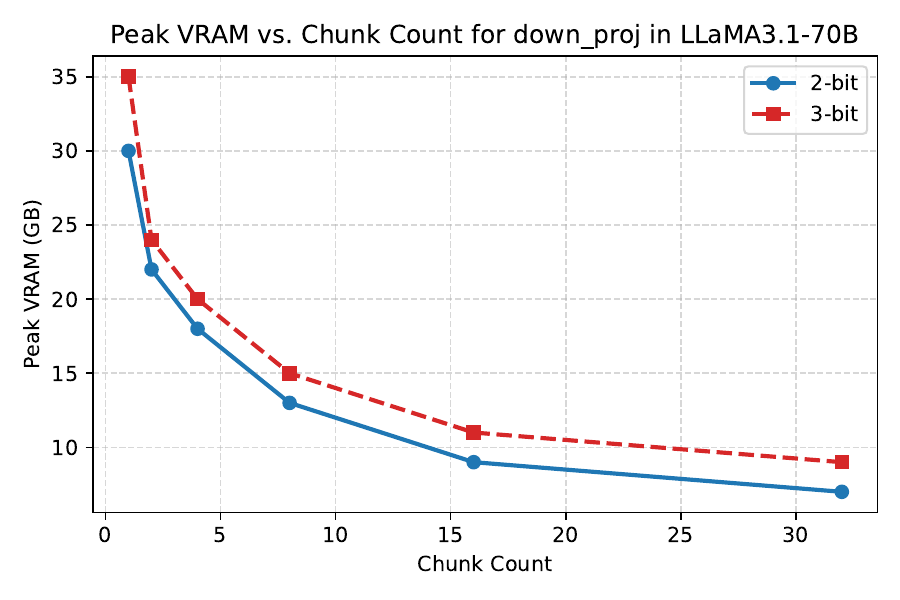} 
\caption{Peak VRAM usage of the Llama-3.1-70B down\_proj matrix under 2-bit and 3-bit quantization with different chunk counts. Increasing the number of chunks significantly reduces the peak memory footprint.}
\label{chunked_computation}
\end{figure}
\subsection{Chunked Computation of HLQ Parameters}
When computing the initial quantization parameters of HLQ for a weight matrix, the optimization within each weight group is independent. For extremely large weight matrices, we can split the matrix into smaller chunks and perform parameter search independently for each chunk. As shown in the figure, for the largest matrix in Llama3.1-70B, the down\_proj of shape [28672, 8192] performing HLQ parameter search without batching leads to a peak memory usage of approximately 30 GB for 2-bit quantization and 35 GB for 3-bit quantization. In contrast, when the search is executed in two chunks, the peak memory drops to about 22 GB for 2-bit and 24 GB for 3-bit quantization. This chunked computation strategy significantly reduces GPU memory consumption and achieves approximately linear memory scaling with respect to partition granularity.

\subsection{Offloading Hidden States to Disk}

During block reconstruction, a large amount of intermediate data (hidden states) must be stored. For example, with a 1K calibration set and a context length of 2048, the hidden size of LLaMA 3.1-8B (hiddensize = 4096) requires about 16 GB of CPU memory in FP16 format, while LLaMA 3.1-70B requires around 64 GB. This imposes heavy memory pressure on the CPU. 

To mitigate this, we offload hidden states to disk. However, fully offloading them leads to excessive disk I/O latency, as disk-to-CPU bandwidth is much lower than CPU-to-GPU bandwidth. To address this, we employ a batched loading strategy: the hidden states are divided into several batches, and only one batch is loaded from disk at a time. For instance, in LLaMA 3.1-8B, we load 4 GB of data ($\frac {1}{4}$ of the total) per iteration. Once a batch is consumed, the next one is fetched from disk. This approach substantially reduces CPU memory usage while minimizing disk access frequency and context switching overhead, thereby ensuring efficient and continuous quantization throughput.
\subsection{Overhead of Disk Access}
Using lazy loading and loading training data directly from disk requires disk access during the execution of quantization algorithms. This inevitably introduces additional memory access overhead. Even with NVMe-based communication protocols, disk bandwidth remains far below that of the CPU. As shown in Table~\ref{tab::disk_and_no_disk_comprasion}, we evaluate  ELUTQ on LLaMA3.1-8B, Qwen3-32B, and LLaMA3.1-70B. The impact of disk access is relatively minor compared to the overall tuning time. Specifically, the tuning time increases by 25\% for LLaMA3.1-8B, 12\% for Qwen3-32B, and only about 5\% for LLaMA3.1-70B. Therefore, when performing ELUTQ on very large models, the overhead caused by disk access can be considered negligible relative to the total tuning time.

\begin{table}[!ht]
\centering
\caption{Comparison of tuning times with and without disk usage. All models are tuned under the same quantization configuration w2g128, using 1k calibration dataset. }
\label{tab::disk_and_no_disk_comprasion}
\begin{tabular}{ccccc}
\toprule
\textbf{Method} & \textbf{Model}  &\textbf{disk(\ding{55})} & \textbf{disk(\ding{51})} &Overhead (\%)  \\
\midrule
\multirow{3}{*}{ELUTQ} &llama3.1-8B & 2.00h & 2.50h & 25\% \\
\cmidrule{2-5}
 & Qwen3-32B & 8.25h & 9.30h & 12\%   \\
 \cmidrule{2-5}
 & llama3.1-70B & 19.00  & 20.00h &  5\% \\ 
\bottomrule
\end{tabular}

\end{table}

\newpage
\section{Design of High Performance CPU and GPU kernels}
\label{cpu_gpu_kernels}
\begin{figure}[t]\rmfamily
\centering
\includegraphics[width=\linewidth]{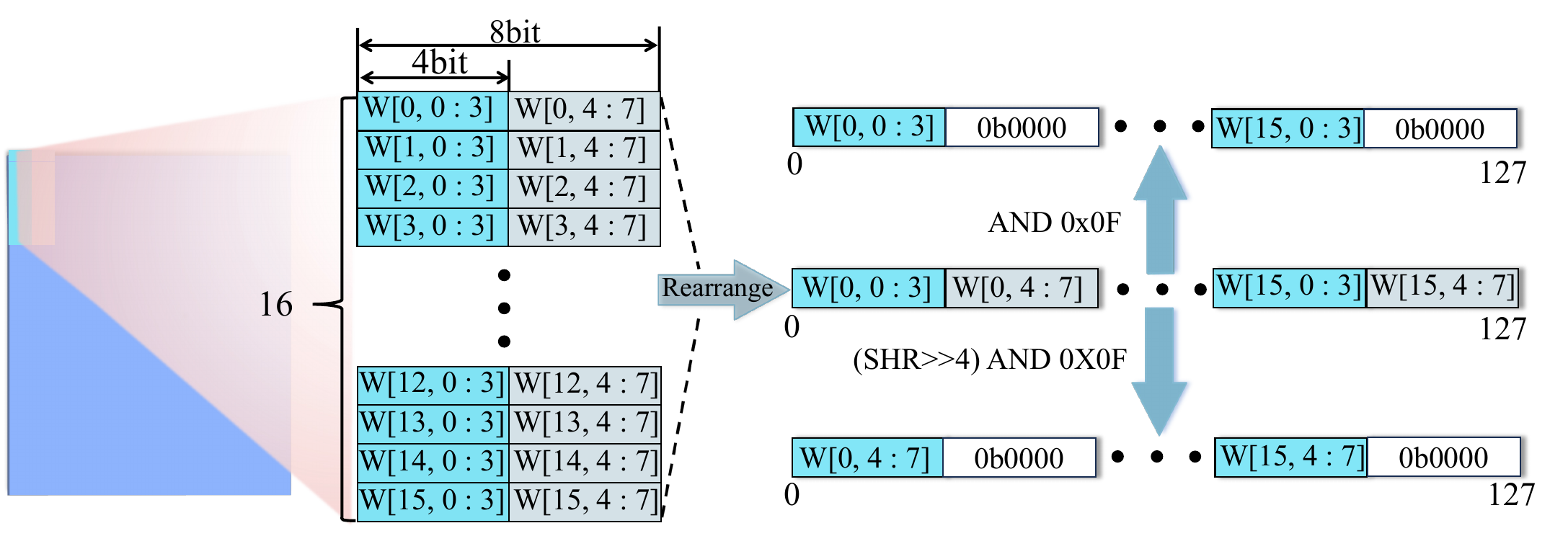} 
\caption{Rearrange weight for ARM NEON's 128-bit registers to ensure memory access continuity and improve decoding speed during runtime.}
\label{weight_rearrange}
\end{figure}

\subsection{Weight-Rearrange.}

Conventional dequantization-based inference frameworks typically store weights in either row-major or column-major order. Some works, such as AWQ, employ interleaved weight storage for 4-bit quantization to accelerate runtime decoding, but the approach fundamentally remains column-major.

Bit-serial LUT-based GEMM operates by loading weights corresponding to activation combinations and performing table lookups to compute results. This requires careful consideration of the weight organization in memory. Taking activation groups of size $g$=4 as an example, Figure~\ref{weight_rearrange} illustrates our design for rearranging the one-bit weight matrics to fit the 128-bit ARM NEON registers and efficiently unpacking them at runtime.

Given ARM’s 128-bit register width, a 16×8 one-bit matrix block stored in conventional row-major or column-major order would reside in non-contiguous memory. To maximize memory bandwidth utilization, we reorganize such blocks along the out-channel dimension, ensuring contiguous memory access. Since weights remain static during inference, this rearrangement can be performed offline, introducing no runtime overhead. During execution, weight decoding only requires simple bitwise AND and shift operations, maintaining computational efficiency.

\subsection{Mirror Storage.}
Given the formulation:
\begin{equation}
\label{equation9}
\hat{\mathrm{W}}=\sum_{j=0}^{q-1}s_j\cdot b_j+z, b_j\in\{0,1\}^n,
\end{equation}
we apply a simple linear transformation by setting $\hat s_j = \frac 1 2 s_j$, $\hat b_j = 2b_j - 1 $, $\hat z = z +  \frac 1 2 \sum_{j=0}^{q-1}s_j$, under this transformation, the quantized weights can be rewritten as:
\begin{equation}
\label{equation9}
\mathrm{\hat W} = \sum_{j=0}^{q-1}\hat s_j\cdot \hat b_j + \hat z,\hat b_j \in \{ -1,1\}^{n}.
\end{equation}
For an input activation combinations of size 4, such as $[x_1,x_2,x_3,x_4]$, the output of the dot product with the weight has 16 possible outcomes, ranging from $(-x_1-x_2-x_3-x_4,\ ...,\ x_1+x_2+x_3+x_4)$. When storing the lookup table, we only need to store half of the possible results, as the remaining half can be obtained by negating the stored values. This table compression method is lossless, fully preserving model inference accuracy while also reducing memory usage by half and accelerating table access.

\subsection{Table Quantization.}
For ARM NEON, activations are typically stored in FP16 precision, and accordingly, each entry in the lookup table also uses FP16 precision. To optimize table storage and access, each FP16 value can be split into two int8 values and interleaved in memory. During table loading, efficient dual-channel load operations such as $vld2q\_u8$ can be employed to construct the table, after which the retrieved values are reconstructed back into FP16.

An alternative approach is to quantize the table itself, mapping FP16 table values to int8. While this may introduce some degradation in model accuracy, it eliminates the need for interleaved storage and FP16 reconstruction, significantly improving runtime efficiency.

\section{Compression Ratios and Actual Memory Footprint of Quantized Models}
\label{compression_ratios}
Here,we provide a comparison of the compression ratios and actual model sizes between hierarchical linear quantization and uniform quantization. Specially, table~\ref{tab::full_compression_rates} show the results of compression radios, while table~\ref{tab::full_actula_model_size} show the results of actual model size. It can be observed that, under the same weight bit-width, HLQ incurs a slightly higher memory footprint compared to uniform quantization.
\begin{table}[ht]
\centering
\caption{Complete comparison of Model Compression Rates ($\uparrow$) between Uniform Quantization and Hierarchical Linear Quantization. Embedding and LM-head layers are excluded from quantization for all models.}
\label{tab::full_compression_rates}
\begin{tabular}{cccccccccc}
\toprule
 \textbf{Wbit}  & \textbf{G}& \textbf{Method} & \textbf{L2-7}  &  \textbf{L2-13} &  \textbf{L3.1-8} & \textbf{L3.1-70} &\textbf{Q3-8} & \textbf{Q3-14} & \textbf{Q3-32} \\
\multirow{2}{*}{2} & \multirow{2}{*}{128} & RTN&5.74 & 6.16 & 3.95 & 6.00 & 3.68  & 4.32 & 5.51\\
 & & HLQ & 5.50 & 5.88  & 3.84 &5.75 & 3.59 & 4.19 & 5.29\\ 
\midrule
\multirow{2}{*}{3} & \multirow{2}{*}{128} & RTN & 4.27 & 4.48 & 3.25 & 4.40 & 3.08 & 3.48 & 4.14 \\
 & & HLQ & 4.01  & 4.19 & 3.11 & 4.13  &2.96 & 3.32  & 3.90\\ 
 \bottomrule
\end{tabular}

\end{table}

\begin{table}[!htbp]
\centering
\caption{Complete comparison of Model size (GB) between Uniform Quantization and Hierarchical Linear Quantization. Embedding and LM-head layers are excluded from quantization for all models.}
\label{tab::full_actula_model_size}
\begin{tabular}{cccccccccc}
\toprule
 \textbf{Wbit}  & \textbf{G}& \textbf{Method} & \textbf{L2-7}  &  \textbf{L2-13} &  \textbf{L3.1-8} & \textbf{L3.1-70} &\textbf{Q3-8} & \textbf{Q3-14} & \textbf{Q3-32} \\
\multirow{2}{*}{2} & \multirow{2}{*}{128} & RTN& 2.18&  3.93 & 3.78 & 21.84 & 4.13 & 6.35 & 11.07 \\
 & & HLQ & 2.27& 4.11 & 3.88 & 22.84 & 4.23 & 6.55 & 11.52\\ 
\midrule
\multirow{2}{*}{3} & \multirow{2}{*}{128} & RTN & 2.93 & 5.41 & 4.59 & 29.81 & 4.95 &7.89  & 14.70\\
 & & HLQ & 3.12  & 5.78 & 4.80 & 31.80& 5.15 & 8.28 & 15.61 \\ 
 \bottomrule
\end{tabular}

\end{table}

\section{Compared with Non-Uniform and Quantization-Aware Training Methods}
\label{non_uniform_and_qat}
In this section, we further compare our results with several codebook-based non-uniform quantization that adopt codebooks and QAT methods. As shown in the table~\ref{tab:compare_with_non_uniform_and_qat}, ELUTQ exhibit a certain gap in model accuracy compared to these non-uniform approaches, which is expected. Methods such as Quip\# \cite{tseng2024quipsharp} and AQLM \cite{egiazarian2024aqlm} require a large calibration set (e.g., 4K training samples with a context length of 4096) and involve complex distillation procedures, leading to substantial computational and time costs during quantization. Meanwhile, when using the same training data, our method achieves comparable results to EfficientQAT \cite{chen2024efficientqat}, however, EfficientQAT necessitates retraining the model weights, whereas our approach operates without any weight training, significantly reducing memory consumption, accelerating quantization, and mitigating the risk of overfitting.

\begin{table*}[ht]
\centering
\caption{Evaluation of quantized LLaMA3.1 models for 2 bits. The table reports perplexity on WikiText2 and c4, as well as accuracy for zero-shot tasks. The Average column is the mean of 5 zero-shot task accuracies.}
\label{tab:compare_with_non_uniform_and_qat}
\resizebox{\textwidth}{!}{
\begin{tabular}{cc|cc|cccccc}
\toprule
\textbf{Model}  & \textbf{Method}&  \textbf{wikitext2} & \textbf{c4}  &  \textbf{WinoGrande} &\textbf{HellaSwag}& \textbf{ArcC} & \textbf{ArcE} & \textbf{PiQA} & \textbf{Average}($\uparrow$)  \\
\midrule
\multirow{6}{*}{LlaMA3.1-8B} & Baseline & 6.15 &  8.89  & 73.32  & 60.04  & 51.28 & 81.57 & 80.03 & 69.25 \\
\cdashline{2-10}
& Quip\# & 11.81 & 14.24 & 65.78 & 51.18 & 38.12 & 70.56 & 76.75  & 60.48 \\
& AQLM & 11.67 & 13.56 & 65.92 & 51.92 & 38.84 & 70.48 & 76.12 & 60.66 \\
& EfficientQAT & 13.12 & 14.95& 64.17 & 49.31 & 34.22 & 69.23 & 74.43 & 58.27 \\
& ELUTQ &12.38 & 15.08 & 62.72 & 49.52 & 32.91 & 69.02 &74.78 & 57.79  \\
\midrule
\multirow{6}{*}{LlaMA3.1-70B} & Baseline & 2.93& 6.92  & 80.72& 66.58 &60.62 &86.99& 82.77& 75.52 \\
\cdashline{2-10}
& Quip\# & 6.89 &10.47 & 73.11 & 60.94 & 50.45 & 77.05 & 78.32 & 67.98  \\
& AQLM & 6.38  &  9.73& 73.56 &61.87 &50.21 &78.26 &79.15 & 68.61\\
& EfficientQAT & 7.35 & 10.56 & 69.27 & 59.95 & 49.04 & 78.84 & 79.22 & 67.26  \\
& ELUTQ & 7.08& 10.65 &  72.38 &  58.86 & 46.52 &77.16 &77.54 &66.70 \\

\bottomrule
\end{tabular}
}

\end{table*}

\section{More details of Efficiency Evaluation}
\label{experiment_configurations_on_memory_time}
\subsection{Experimental Configurations for Memory and Time Comparison}
In this section, we provide the detailed experimental configurations used to measure the computational memory consumption (CPU and GPU) and quantization time of different quantization methods on LLaMA3.1–70B under 2-bit quantization. All the experiments is conducted on RTX 4090s.

\begin{itemize}
    \item GPTQ: Following the official implementation, we use 128 calibration samples with a context length of 2048 and perform layer-wise quantization, using 2 RTX4090s.

\item AWQ: According to the official codebase, we use 512 calibration samples from the default AWQ dataset, each with a context length not exceeding 512, using single RTX4090.

\item OmniQuant: We use 128 calibration samples with a context length of 2048 and follow the official recommendation to iterate 20 epochs for each transformer block, using 2 RTX4090s.

\item Quip\#: We use 4K calibration samples with a context length of 4096 and conduct both layer-wise quantization, using 4 A100-80GB GPUs.

\item AQLM: We follow the same configuration as Quip\#, using 4K calibration samples with a context length of 4096, and perform block-wise quantization,using 4 A100-80GB GPUs.

\item DB-LLM: We use a self-collected dataset of 20K samples for calibration, using 4 A100-80GB GPUs.

\item EfficientQAT: We use 4K calibration samples with a context length of 2048,iterate 2 epochs for each transformer block, using 2 RTX4090s in block-wise reconstruction and 2 RTX4090s in end to end tuning.

\item ELUTQ: We use 4K calibration samples with a context length of 2048,iterate 2 epochs for each transformer block, using single RTX4090 in block-wise reconstruction and 2 RTX4090s in end to end tuning.
\end{itemize}

\subsection{Parameter Efficiency}
ELUTQ updates only a small subset of parameters in both stages, making it highly parameter-efficient. For LLaMA3.1-8B, we conduct experiments comparing the number of trainable parameters required for fine-tuning a single Transformer block and the entire model under different strategies, as shown in table~\ref{tab:parameter_efficiency}. Our method is parameter-efficient.

\begin{table}[ht]
\centering
\caption{Comparison of the number of trainable parameters required by different fine-tuning strategies for w2g128 LLaMA3.1-8B.}
\label{tab:parameter_efficiency}

\begin{tabular}{c|cc}
\toprule
\textbf{Strategy} & \textbf{Block-level}  & \textbf{Model-level} \\
\midrule

W&  218M  & 7G  \\
s,z,round & 221M & 7.1G \\
clipping & 3M & 0.1G \\
round & 218M & 7G \\
s,z,W & 221M & 7.1G \\
ELUTQ(ours) & 5M & 0.15G \\ 
\bottomrule
\end{tabular}

\end{table}

\newpage
\section{Full Results of Inference Performance on Edge Devices}
\label{Full_results_on_edge_devices}
\subsection{Evaluation on CPUs}
\label{overhead_of_HLQ}
We modified the T-MAC operators to support the HLQ format and evaluated their performance at the kernel level. Table~\ref{overhead of HLQ} presents the latency results of two types of matrix operations in the 2-bit quantized LLaMA2-7B model, evaluated on both the RK3588 and Apple M2 chips. For each operation, the reported latency is the average of 10 runs. The results demonstrate that the introduction of HLQ does not introduce additional overhead to the Bit-serial LUT-based GEMM. The increase in average latency is negligible compared to the inherent fluctuations and overall runtime. Therefore, although the HLQ quantization format occupies slightly more memory than uniform quantization under the same bit-width and group configuration, it does not introduce any additional computational overhead during inference. Moreover, the extra memory required by HLQ is small relative to the overall size of the compressed model, typically remaining below 5\%, which is considered acceptable in practice.

\begin{table}[ht]
\centering
\caption{Comparison of GEMV latency (ms) between uniform quantization and HLQ formats using the T-MAC kernel.}
\label{overhead of HLQ}
{\fontsize{9}{11}\selectfont
\setlength{\tabcolsep}{3pt} 
\begin{tabular}{c|c|c|c}
\toprule
Chip & Kernel  & \textbf{$ 11008 \times 4096$} & \textbf{ $4096 \times 32000$} \\
\midrule
\multirow{2}{*}{RK3588}& T-MAC-Uniform   &  $5.70\,(\pm\,0.10)$  & $22.11\,(\pm\,0.17)$ \\

   & T-MAC-HLQ       & $5.72\,(\pm\,0.11)$ &   $22.23\,(\pm\,0.16)$ \\
\midrule

 \multirow{2}{*}{M2}& T-MAC-Uniform       & $1.84\,(\pm\,0.07)$   & $4.46\,(\pm\,0.05)$    \\
 & T-MAC-HLQ & $1.86\,(\pm\,0.04)$   &  $4.49\,(\pm\,0.05)$ \\
\bottomrule
\end{tabular}
}

\end{table}

\begin{table}[ht]
\centering
\caption{End-to-end performance comparison of ELUTQ and AWQ on LLaMA3.1-8B running on Apple M2 (batch size = 1, prompt length = 512).}
\label{tab:compared with awq}

\begin{tabular}{ccccc}
\toprule
\textbf{Method} & \textbf{W}  & \textbf{G} & \textbf{Mem(GB)} &\textbf{Speedup}\\
\midrule
Baseline& 16    &  - & 15.0 & 1.0$\times$ \\
\midrule

AWQ& 3    & 128 & 4.7& 2.0$\times$ \\
\textbf{ELUTQ}& 3       & 128  & 4.8 & \textbf{2.5$\times$}\\
\midrule
AWQ& 2    & 128 & 3.9& 1.6$\times$ \\
\textbf{ELUTQ} & 2       & 128  & 3.9 & \textbf{3.4$\times$} \\
\bottomrule
\end{tabular}

\end{table}

\begin{table*}[htbp]
\centering
\caption{Comparison of prefill and decode throughput (tokens/s) between ELUTQ and llama.cpp on the Apple M2 chip. All experiments are conducted with 4 threads and a batch size of 1.}
\label{tab:performance_comparison}
\setlength{\tabcolsep}{6pt} 
\begin{tabular}{ccccccc}
\toprule

\textbf{Model} & \textbf{Framework}& \textbf{Wbits} &\textbf{BPW} & \textbf{Input,Output} & \textbf{Prefill} & \textbf{Decode} \\ 
\midrule
\multirow{10}{*}{llama2-7B} & \multirow{6}{*}{llama.cpp} & \multirow{2}{*}{Q3\_K\_M} & \multirow{2}{*}{3.91} & 128,128 & 19.87 & 15.65 \\
 &  &  & & 256,128 & 19.75 & 15.53 \\
 & & \multirow{2}{*}{Q3\_K\_S} &\multirow{2}{*}{3.50}& 128,128 & 17.60 & 15.44  \\
 & &  & & 256,128 & 17.43 & 15.00   \\
 & & \multirow{2}{*}{Q2\_K\_S} & \multirow{2}{*}{2.50}&128,128 & 14.82& 13.35\\
 & &  & &  256,128  & 14.54& 13.16 \\
\cmidrule{2-7}
 & \multirow{4}{*}{ELUTQ}& \multirow{2}{*}{W3(g128)} & \multirow{2}{*}{3.50} & 128,128 & 20.60 & 16.63   \\
 & &  &  & 256,128 & 21.54 & 16.21  \\
 & & \multirow{2}{*}{W2(g128)} & \multirow{2}{*}{2.37} & 128,128  & 25.06& 20.48  \\
  & &  &  & 256,128 &25.94 & 20.25  \\ 
\bottomrule
\end{tabular}

\end{table*}

\label{more_results_on_edge_devices}
\begin{table*}[!htbp]
\centering
\caption{Comparison of prefill and decode throughput (tokens/s) between ELUTQ and llama.cpp on the Apple M2 chip and RK3588. All experiments are conducted with 4 threads and a batch size of 1.}
\label{tab:more_results_on_edge_devices}
\setlength{\tabcolsep}{6pt} 
\begin{tabular}{ccccccccc}
\toprule

\multirow{2}{*}{\textbf{Model}} & \multirow{2}{*}{\textbf{Framework}} & \multirow{2}{*}{\textbf{Wbits}} & \multirow{2}{*}{\textbf{BPW}} & \multirow{2}{*}{\textbf{Input,Output}} & \multicolumn{2}{c}{\textbf{Apple M2}} & \multicolumn{2}{c}{\textbf{RK3588}} \\
 & & & & & Prefill & Decode & Prefill & Decode\\ 
\midrule
\multirow{10}{*}{Qwen2-1.5B} & \multirow{6}{*}{llama.cpp} & \multirow{2}{*}{Q3\_K\_M} &\multirow{2}{*}{3.91} & 128,128 & 103.11 & 57.69 & 21.53 & 15.36 \\
 &  &  &  & 256,128 & 101.37 & 56.11  & 21.28 & 15.13\\
 & & \multirow{2}{*}{Q3\_K\_S} & \multirow{2}{*}{3.50} & 128,128 & 86.67 & 54.64 & 19.79 & 14.57\\
 & &  &  & 256,128 & 85.54 & 53.49 & 19.58 & 14.37\\
  & & \multirow{2}{*}{Q2\_K\_S} & \multirow{2}{*}{2.50} & 128,128  &  62.73&  48.55  &  16.12 &  12.95 \\
  & &  &  & 256,128 & 60.17  & 47.87 & 15.98 & 12.78 \\
 \cmidrule{2-9}
 &\multirow{4}{*}{ELUTQ} & \multirow{2}{*}{W3(g128)} & \multirow{2}{*}{3.50}& 128,128 & 88.72 & 55.03 &  27.77 & 14.84\\
 & &  &  & 256,128 & 93.06 & 54.90 & 27.16 & 14.54 \\
 & & \multirow{2}{*}{W2(g128)} & \multirow{2}{*}{2.37} & 128,128 & 120.65& 65.19 &  33.61 & 17.42 \\
  & &  &  & 256,128 & 124.36  & 64.81 &  35.39 &  17.15\\
\bottomrule
\end{tabular}

\end{table*}

\newpage
\subsection{Evaluation on GPUs}
\label{results_on_gpus}
Here, we provide the experimental results on GPUs. Our evaluation covers a range of hardware platforms, including high-performance server-grade chips (A100 and H100),  consumer-grade GPUs (RTX 4090 and RTX 3090). As baselines, we consider AWQ, which uses a uniform quantization format, and SqueezeLLM, which adopts a non-uniform quantization format. We convert our HLQ weights into the BCQ format via a simple linear transformation and perform inference using the AnyBCQ framework. The results are presented in the table~\ref{tab:results_on_gpus}.
\begin{table*}[!htbp]
\centering
\caption{End to End Evaluation (tokens/s) of generating a sequence lengeth of 1024 on GPUs. OOM indicates that the model cannot be deployed on the corresponding GPU.}
\label{tab:results_on_gpus}
\begin{tabular}{ccccccc}
\toprule
\textbf{Model}  & \textbf{Wbits}&  \textbf{Method} & \textbf{A100}  &  \textbf{H100} &\textbf{RTX3090}& \textbf{RTX4090}  \\
\midrule
\multirow{7}{*}{LLaMA3.1-8B} & 16 & Baseline & 100 & 165 & 55 & 60 \\
\cmidrule{2-7}
 & \multirow{3}{*}{2} & AWQ &  175 & 217 & 119 & 138\\
& & SqueezeLLM & 160 & 189 & 99 & 109\\
& & BCQ & \textbf{256} & \textbf{311}  & \textbf{184} & \textbf{210} \\
\cmidrule{2-7}
&  \multirow{3}{*}{3} & AWQ & 198 & 233 & 143 & 164 \\ 
& & SqueezeLLM & 185 & 228 & 130 & 141 \\
& & BCQ &  \textbf{220} & \textbf{282} & \textbf{155} & \textbf{174} \\
\midrule
\multirow{7}{*}{LLaMA2-13B} & 16 & Baseline & 60 & 105 & OOM & OOM \\
\cmidrule{2-7}
 & \multirow{3}{*}{2} & AWQ & 104 & 146 & 79 & 91 \\
& & SqueezeLLM & 92 & 107 & 75 & 80\\
& & BCQ & \textbf{185} & \textbf{246} & \textbf{135} & \textbf{152} \\
\cmidrule{2-7}
&  \multirow{3}{*}{3} & AWQ & 123 & 178 & 88 & 104\\ 
& & SqueezeLLM & 10 & 155 & 80 & 87\\
& & BCQ & \textbf{152} & \textbf{208} & \textbf{106} & \textbf{118} \\
\midrule
\multirow{7}{*}{LlaMA3.1-70B} & 16 & Baseline & OOM & OOM & OOM & OOM \\
\cmidrule{2-7}
 & \multirow{3}{*}{2} & AWQ & 28 & 44 & OOM & OOM \\
& & SqueezeLLM & 22 & 38 & OOM & OOM \\
& & BCQ & \textbf{55} & \textbf{98} & OOM & OOM \\
\cmidrule{2-7}
&  \multirow{3}{*}{3} & AWQ & 34 & 54 & OOM & OOM\\ 
& & SqueezeLLM & 28 & 45 & OOM & OOM \\
& & BCQ & \textbf{43} & \textbf{79} & OOM & OOM \\
\bottomrule
\end{tabular}

\end{table*}


\newpage
\newpage
\section{Full Results on Model Accuracy}
\label{full_results}
Here, we present the full results of our method across various models. Specifically, Table~\ref{tab:wikitext2_ppl_results} reports the perplexity on wikitext2, Table~\ref{tab:llama_zeroshots_results} reports the accuracy on the LLaMA series models, while Table~\ref{tab:qwen_zeroshots_results} shows the accuracy on the Qwen series models.

\begin{table*}[!htbp]
\centering
\caption{A comparison of wikitext2 perplexity ($\downarrow$) between weight-only quantization methods, with a context length of 2048. \textbf{BPW} represents the average number of bits per weight. Scale and zero-point are assumed to be stored in fp16 format. PB-LLM* denotes the result of PB-LLM with GPTQ and 20\% salient weight. "–" indicates that the framework does not support this model. }
\label{tab:wikitext2_ppl_results}
\begin{tabular}{ccccccccccc}
\toprule
\textbf{Method} & \textbf{\# W}  & \textbf{\# G}&  \textbf{BPW} & \textbf{L2-7}  &  \textbf{L2-13} &  \textbf{L3.1-8} & \textbf{L3.1-70} &\textbf{Q3-8} & \textbf{Q3-14} & \textbf{Q3-32} \\
\midrule
Baseline & 16 & - & 16   &5.47   & 4.88  & 6.15 & 2.81 &  9.72 & 8.64 & 7.61\\
\midrule
GPTQ         & 2 & 128  & 2.25     & 16.01  & 10.33   & 109.30  &22.54 & 29.19 & 26.15 &  23.18 \\
AWQ         & 2 &128  & 2.25  & 2.21e5  & 1.23e5  & 1.74e6 & 1.32e6& 1.21e5 & 1.42e5 & 1.43e4\\
OmniQuant & 2   &128  & 2.25   & 11.06  &  8.26  & 19.18 & 16.21 & 24.67 & 19.41 & 15.74\\
ApiQ & 2 & 128 & 2.25 & 8.25 & 6.84 & 13.18 & - & - & - & - \\
PB-LLM* & - & -  & 2.2    & 17.19  & 12.47 & 21.84  & 18.72 & - & - & -\\
ShiftAddLLM & 2  & 128 & 2.37 &  9.58 & 12.57 &  34.41 & 12.46 & - & - & - \\
AnyBCQ & 2  & 128 & 2.37 & 8.96 & 7.14 & 19.01 & 10.17 & - & - & - \\

\textbf{ELUTQ} & 2  &  128 &  2.37  & \textbf{8.05}  & \textbf{6.75} &\textbf{12.38} &\textbf{7.08} & \textbf{17.87} & \textbf{14.92} & \textbf{10.31} \\ 
\midrule
DB-LLM & 2   &64 &  -  & 7.23   & 6.19  & 13.11 & - & - & - & - \\

\textbf{ELUTQ} & 2  &64 &  2.75   &  \textbf{7.07}   &  \textbf{6.07} &\textbf{9.72} & \textbf{6.56}& \textbf{15.09} & \textbf{12.24} &\textbf{9.72}  \\

\midrule
GPTQ & 3  &128 & 3.25     &6.29  & 5.42& 9.58&  5.67 & 10.76 & 9.69 & 9.15 \\ 
AWQ  & 3  &128 & 3.25  & 6.24  & 5.32   & 8.22 &  5.51 & 14.90 & 10.41 & 8.89\\
OmniQ& 3  &128 & 3.25  & 6.03  & 5.28 & 8.27 & 5.66 & 11.71 & 9.86  & 9.04\\ 
SqueezeLLM(0.45\%) & 3 & - & 3.25 & 5.96 & 5.23 & 7.48 & - & - & - & - \\
GANQ & 3 & - & 3.25 & 5.96 & 5.20 & 7.46 & - & - & - & - \\

ShiftAddLLM & 3  & 128 & 3.50 &   6.04 & 5.33 &  7.71 & 4.66 & - & - & -\\

AnyBCQ & 3  & 128 & 3.50 &   6.05& 5.28 &  7.68 &  4.47  & - & - & -\\
\textbf{ELUTQ} &  3& 128 & 3.50 & \textbf{5.93} & \textbf{5.12} & \textbf{7.01}& \textbf{4.22} & \textbf{9.95}  & \textbf{9.21} & \textbf{7.92}\\ 
\bottomrule
\end{tabular}

\end{table*}

\newpage

\begin{table*}[!htbp]
\centering
\caption{ Accuracy (\%) on zero-shot tasks on LlaMA models by lm\_eval v0.4.9. PB-LLM* denotes the result of PB-LLM with GPTQ and 20\% salient weight. \textbf{Bold}: best result; \underline{underlined}: second-best. "–" indicates that the framework does not support this model. Acc is reported, not acc norm. }
\label{tab:llama_zeroshots_results}
\resizebox{\textwidth}{!}{
\begin{tabular}{cccccccccc}
\toprule
\textbf{Model} & \textbf{Method}  & \textbf{Bits} & \textbf{Group} & \textbf{WinoGrande} &\textbf{HellaSwag}& \textbf{ArcC} & \textbf{ArcE} & \textbf{PiQA} & \textbf{Average}($\uparrow$)\\
\midrule
\multirow{12}{*}{LLaMA2-7B} &  Baseline  & 16& - &  70.24 & 56.86 & 41.04 & 74.66 & 77.86 & 64.13  \\
\cdashline{2-10}
        & GPTQ     & 3  & 128& 66.93&54.55& 38.31 & 70.54 & 77.04 & 61.47  \\
        & AWQ      & 3  & 128 & 69.77 & 54.73 & 38.82 & 71.59 &  76.22 & 62.23  \\
    & OmniQ       & 3  & 128 & 66.19 & 54.12 &38.72 &  72.15 & 77.12&61.66  \\
    & \textbf{ELUTQ} & 3 & 128 & 69.30 & 54.72 &  39.42 & 72.98 & 77.16 & \textbf{62.72}\\
\cmidrule{2-10}
    & GPTQ      &2  & 128 & 53.12 & 34.17 & 21.33  & 35.40 & 59.30 & 40.66 \\
        & AWQ      & 2  & 128 & 49.72  & 25.38 & 22.53 & 25.46 & 52.67 & 35.15 \\ 
    & OmniQ       & 2  & 128 &55.88 &40.28 &23.46 &50.13& 65.13 & 46.98 \\
    & PB-LLM*      & -  & - & 50.36& 30.49 &22.01 & 29.88 &55.22 & 37.60 \\
    & \textbf{ELUTQ}     & 2 & 128  & 62.75 &46.61 &30.72 &63.01 &72.25& \textbf{55.07}  \\
\midrule
\multirow{11}{*}{LLaMA2-13B} &  Baseline  & 16 & -& 72.22 &60.07 &48.29 &79.42& 79.05 & 67.81 \\
\cdashline{2-10}
        & GPTQ     & 3  & 128 & 70.88 & 57.83 & 45.65& 77.99 &78.56 &66.18 \\
        & AWQ       & 3  & 128 &  71.82 & 58.58 & 44.62 & 77.95& 77.75&  66.14\\
    & OmniQ       & 3 & 128 &  70.01 & 58.46 &46.16 &77.86 &78.40 &66.18 \\
    & \textbf{ELUTQ} & 3 & 128 &  71.94 &59.54 &46.32 &78.70 &78.56& \textbf{67.02}\\
\cmidrule{2-10}
 & GPTQ & 2 & 128 & 55.80 &41.06 &21.93 &55.60 &67.08 &48.29\\
 &OmniQ & 2 & 128  & 57.93 &46.23 &30.29 &63.22 &70.13 &53.56   \\
 & PB-LLM* & - & - &   52.33 &  30.23 &  23.12  & 31.27 & 55.01 & 38.39 \\
  & \textbf{ELUTQ} & 2 &  128 &  66.38 &50.78 &36.18& 67.72 &75.19 &\textbf{59.25} \\        
\midrule
\multirow{8}{*}{LLaMA3.1-8B} &  Baseline  &16 & - & 73.32 &60.04 &51.28 &81.57 &80.03 & 69.25  \\
\cdashline{2-10}
        & GPTQ      &3  & 128& 70.72 &55.58 & 42.75  &74.54&77.48 &64.21 \\
        & AWQ      & 3  & 128 & 70.96 & 55.41 & 44.45 & 76.53 & 77.58 & 64.98  \\ 
    & \textbf{ELUTQ} & 3 & 128 & 71.25 &55.75& 45.65& 77.31 &78.13 & \textbf{65.62} \\
\cmidrule{2-10}
&GPTQ  & 2 & 128 & 48.78 & 27.43 & 19.20 & 28.96 & 55.55 & 35.98  \\
  &\textbf{ELUTQ} & 2  & 128  & 62.72 & 49.52 & 32.91 & 69.02 &74.78 &  \textbf{57.79} \\
\midrule
\multirow{8}{*}{LLaMA3.1-70B} &  Baseline  &16 & - & 80.72& 66.58 &60.62 &86.99& 82.77& 75.52 \\
\cdashline{2-10}
        & GPTQ      &3  & 128 & 77.94& 62.76 &  52.59 &  81.74  &80.50 & 71.08 \\
        & AWQ      & 3  & 128 &  78.13 &63.51 &56.01 &82.71 &80.28 & 72.13 \\ 
    & \textbf{ELUTQ} & 3 & 128 & 79.48 & 64.92 & 56.91 & 83.25 & 81.10 & \textbf{73.13} \\
\cmidrule{2-10}
&GPTQ  & 2 & 128 & 59.92 & 41.07 & 24.65 & 38.95 & 62.78 & 45.57  \\
  &\textbf{ELUTQ} & 2  & 128  & 72.38 & 58.86 & 46.52 & 78.16 & 77.54 & \textbf{66.70} \\
\bottomrule
\end{tabular}
}

\end{table*}

\newpage
\begin{table*}[ht]
\centering
\caption{ Accuracy (\%) on zero-shot tasks on Qwen models by lm\_eval v0.4.9. \textbf{Bold}: best result; \underline{underlined}: second-best. "–" indicates that the framework does not support this model. Acc is reported, not acc norm.}
\label{tab:qwen_zeroshots_results}
\resizebox{\textwidth}{!}{
\begin{tabular}{cccccccccc}
\toprule
\textbf{Model} & \textbf{Method}  & \textbf{Bits} & \textbf{Group} & \textbf{WinoGrande} &\textbf{HellaSwag}& \textbf{ArcC} & \textbf{ArcE} & \textbf{PiQA} & \textbf{Average}($\uparrow$)\\
\midrule
\multirow{9}{*}{Qwen3-8B} &  Baseline  & 16& - &  67.80 & 57.11 & 55.55 & 83.42 & 76.61 & 68.10   \\
\cdashline{2-10}
        & GPTQ     & 3  & 128& 65.75 & 53.83 & 44.80 & 75.55 & 75.24 & 63.04  \\
        & AWQ      & 3  & 128 & 62.25 & 54.71 & 41.03 & 68.27 & 71.98 & 59.65 \\
    & \textbf{ELUTQ} & 3 & 128 & 68.27 & 53.88 & 52.13 & 80.98 & 76.77 & \textbf{66.41}\\
\cmidrule{2-10}
    & GPTQ      &2  & 128 & 51.48 & 30.45 & 21.56 & 37.82 & 55.92 & 39.45 \\
        & AWQ      & 2  & 128 & 51.72 & 25.32 & 23.81 & 25.76 & 52.47 & 35.82 \\ 
    & \textbf{ELUTQ}     & 2 & 128  &  64.56 & 46.07 & 41.21 & 72.18 & 73.18 & \textbf{59.44} \\

\midrule
\multirow{9}{*}{Qwen3-14B} &  Baseline  & 16& - &  72.93 & 60.93 & 58.53 & 84.13 & 79.92 & 71.29   \\
\cdashline{2-10}
        & GPTQ     & 3  & 128&  70.22 & 57.33 & 55.10 & 77.49 & 76.26 & 67.28 \\
        & AWQ      & 3  & 128 & 57.06 & 55.64 & 48.81 & 75.70 & 74.32 & 62.31  \\
    & \textbf{ELUTQ} & 3 & 128 &  72.56 & 59.50 & 56.31 & 82.74 & 79.05 & \textbf{70.03}\\
\cmidrule{2-10}
    & GPTQ      &2  & 128 & 58.74 & 35.12 & 31.87 & 51.26 & 67.79 & 48.96 \\
        & AWQ      & 2  & 128 & 53.87 & 28.16 & 27.31  & 29.16 & 58.78 & 39.46   \\ 
    & \textbf{ELUTQ}     & 2 & 128  & 70.96 & 50.54 & 48.46 & 79.59 & 75.41 & \textbf{64.99}  \\

\midrule
\multirow{9}{*}{Qwen3-32B} &  Baseline  & 16& - &  73.65 & 63.92 & 57.85 & 84.42 & 80.93 & 72.15   \\
\cdashline{2-10}
        & GPTQ     & 3  & 128& 72.21 & 56.17 & 50.92 & 78.60 & 74.31 & 66.44  \\
        & AWQ      & 3  & 128 & 72.45 & 57.64 & 50.87 & 79.31 & 75.25 & 67.10 \\
    & \textbf{ELUTQ} & 3 & 128 & 72.76 & 61.57 & 55.45 & 82.23 & 79.03 & \textbf{70.21}\\
\cmidrule{2-10}
    & GPTQ      &2  & 128 & 56.62 & 40.67 & 35.92 & 61.30 & 57.19& 50.34 \\
        & AWQ      & 2  & 128 & 53.56 & 35.79 & 32.43 & 58.27 & 53.65 & 46.76\\ 
    & \textbf{ELUTQ}     & 2 & 128  & 70.98 & 55.02 & 48.45 & 78.48 & 74.92 &  \textbf{65.57} \\
\bottomrule
\end{tabular}
}

\end{table*}
\end{document}